\documentclass[runningheads]{llncs}

\usepackage[T1]{fontenc}
\usepackage{marvosym}
\usepackage{graphicx}
\usepackage{amsmath}

\usepackage{amsthm}
\usepackage[section]{placeins}
\usepackage{booktabs}
\usepackage{algorithm}
\usepackage{algorithmic}
\usepackage{bbm}
\usepackage{xcolor}
\usepackage{subfigure}
\usepackage{bm}
\usepackage{amssymb}

\begin{document}
\title{Multi-agent Multi-armed Bandits with Stochastic Sharable Arm Capacities}
%
%
\author{
	Hong Xie\inst{1,2}\textsuperscript{(\Letter)}  
	\and 
	Jinyu Mo\inst{4}
	\and 
	Defu Lian\inst{1,2}
	\and 
	Jie Wang\inst{3}
	\and 
	Enhong Chen\inst{1,2}
}

\authorrunning{H. Xie et al.}
%
\institute{School of Computer Science and Technology \\
University of Science and Technology of China\\ 
	\and
	State Key Laboratory of Cognitive Intelligence, China 
	\and 
	Department of Electronic Engineering and Information Science \\
	University of Science and Technology of China
	\and
	College of Computer Science, Chongqing University, China 
}
\maketitle              
\begin{abstract}
Motivated by distributed selection problems, 
we formulate a new variant of multi-player 
multi-armed bandit (MAB) model, 
which captures stochastic arrival of requests to each arm, 
as well as the policy of allocating requests to players.  
The challenge is how to design a distributed learning 
algorithm such that players select arms according to the optimal arm pulling profile 
(an arm pulling profile prescribes the number of players at each arm) 
without communicating to each other.   
We first design a greedy algorithm, which locates one of the optimal 
arm pulling profiles with a polynomial computational complexity.     
We also design an iterative distributed algorithm for players 
to commit to an optimal arm pulling profile 
with a constant number of rounds in expectation.   
We apply the explore then commit (ETC) framework 
to address the online setting when model parameters are unknown.  
We design an exploration strategy 
for players to estimate the optimal arm pulling profile.   
Since such estimates can be different across different players, 
it is challenging for players to commit.  
We then design an iterative distributed algorithm, 
which guarantees that players can arrive at a consensus on the 
optimal arm pulling profile in only $M$ rounds.  
We conduct experiments to validate our algorithm.    
\end{abstract}
\section{Introduction}
Over the past decade, multi-player MAB has attracted extensive attentions 
\cite{Awerbuch2008,Anandkumar2011,Besson2018,Boursier2019,Lugosi2018,Rosenski2016,Wang2020}.     
The canonical multi-player MAB model \cite{Anandkumar2011,Besson2018}
was motivated from the channel access problem in cognitive radio applications.  
In this channel access problem, 
multiple secondary users (modeled as players) 
compete for multiple channels (modeled as arms).  
In each decision round, each player can select one arm.  
When collision happens (i.e., multiple players selecting the same arm), 
all players in the collision receive no reward.  
Players can not communicate with each other 
and they are aware of whether they encounter a collision or not.  
The objective is to maximize the total reward of all players.  
A number of algorithms were proposed  
\cite{Anandkumar2011,Besson2018,Boursier2019,Lugosi2018,Rosenski2016,Wang2020}.    
Various extensions of the canonical model were studied 
\cite{Bubeck2020,Boursier2019,Hanawal2021,Mehrabian2020,Dubey2020} 
and one can refer to our related work section for more details.  

Existing multi-player MAB models are mainly built on the reward model 
that in each round either 
only one reward is generated from an arm 
\cite{Anandkumar2011,Rosenski2016,Besson2018}, 
or multiple rewards are generated but the 
number of rewards equals the number of players 
\cite{Dubey2020,Landgren2016,Martinez2019}.  
And they are based on the collision assumption that 
when collision happens either all players in the collision receive no reward 
(one reward is generated),  
or each player receives an independent reward.  
The reward model and collision model make 
existing multi-player models not a satisfactory model for solving  
distributed selection problems arise from 
ridesharing applications like 
Uber (\url{https://www.uber.com/},  
food delivery applications like DoorDash 
(\url{https://www.doordash.com/}), etc. 
For those applications, an arm can model a riding pickup location or 
food pickup port.  
A player can model a driver or a delivery driver.  
The ridesharing request or food delivery request 
arrives at an arm in a stochastic manner, 
which is independent of the number of players who will select this arm.  
In case of collision each player can serve 
at most one request in the manner that 
if the number of requests exceeds 
the number of players in the collision, 
each player serves one request and the remaining requests are unserved, 
and on the contrary the remaining players are idle serving no requests. 

To model and design efficient algorithms for allocating requests to players, 
we formulate a new variant of the multi-player MAB model to 
address the above distributed selection problem.   
Our model consists of $M \in \mathbb{N}_+$ arms and $K  \in \mathbb{N}_+$ players.  
For each arm, the number of requests across different decision rounds 
are independent and identically distributed (IID) 
samples from a probability distribution (called request distribution)
and the reward of different requests are 
IID samples from another probability distribution (called reward distribution).  
The request distribution and reward distribution across different 
arms can be different.  
Each player is allowed to serve at most one request.  
When the number of requests in an arm exceeds 
the number of players who pull the arm, 
each player can serve only one request and the remaining requests are unserved, 
and on the contrary,  
the remaining players will be idle serving no requests. 
In the request assigning process, 
there is no differentiation between players or between requests.  
Players can not communicate with each other.  
When a decision round ends, the platform makes the number of requests  
and the number of players on each arm public to all players.  
The objective is to maximize the total reward of all players without knowing 
the request distribution and reward distribution.  
To illustrate: 

\begin{example}
Consider $M=3$ arms and $K=2$ players.  
Each arm can be a pickup location and each player can be a driver. 
For simplicity, in each time slot, 
two/two/one ride-sharing requests arrive at arm 1/2/3.  
Each request in arm 1/2/3 is associated 
with a reward of 0.2/0.2/0.3.    
Let $n_{t,1}, n_{t,2}, n_{t,3}$ denote the number of players 
who pull arm (or go to pickup location) 1, 2 and 3 respective.  
All possible arm pulling profiles, i.e., $(n_{t,1}, n_{t,2}, n_{t,3})$, 
can be expressed as $\{ (2,0,0), (0,2,0), (0,0,2), (1,1,0), (0,1,1),(1,0,1) \}$.  
The total number of arm pulling profiles can 
be calculated as ${M+K-1 \choose M-1} = {4 \choose 2} = 6$.   
Among them, the arm pulling profiles (0,1,1) and (1,0,1) 
achieve the highest total reward, i.e., $0.3+0.2 = 0.5$.  
Players can not communicate with each other in pulling arms.  
In the first time slot, the arm pulling profile can be 
$(0,0,2)$, i.e., both players pull arm 3.  
Then, one player serves a request and receive a reward of 0.3, 
and the other one gets no requests (the allocation of 
requests to players is specified by the application itself).  
In the second time slot, 
the arm pulling profile can be 
$(1,1,0)$,   
then each player gets a reward of 0.2.     
\label{exp:Intro}
\end{example}

\noindent
Example \ref{exp:Intro} illustrates that under the simplified setting where 
reward and arrival are deterministic and given, 
the number of all possible arm pulling profiles is ${M+K-1 \choose M-1}$.  
Exhaustive search is computationally infeasible when the number of 
arms and players are large.   
In practice, both reward and arrival are stochastic 
capturing uncertainty in real-world applications, 
which further complicates the problem.   
\textit{How to design computationally efficient searching algorithms  
to locate the optimal arm pulling profile under the offline setting?}   
Once the searching algorithm is developed, 
players can use it locally to locate an optimal arm pulling profile.  
As illustrated in Example \ref{exp:Intro}, 
there can be multiple optimal arm pulling profiles.   
Players cannot communicate with each other on which one they should commit to.  
\textit{How to design a distributed algorithm 
such that players will commit to an optimal arm pulling profile?}   
After we address the aforementioned two challenges,  
we turn our attention to address the online setting via the explore then commit (ETC) framework, 
which was also used in previous works \cite{Bistritz2018,Rosenski2016}.  
In the exploration phase, players estimate the request distribution 
and reward distribution.    
Using these estimates of distributions in the searching algorithm developed 
in the offline setting, each player can obtain an 
estimate of the optimal arm pulling profile.  
When there are multiple optimal arm pulling profiles, 
these estimates may classify some optimal arm pulling profiles 
as suboptimal ones.  
Different players can have different estimates of distributions and therefore have different estimates 
of arm pulling profiles.   
What makes it challenging is that the estimate of the optimal arm pulling profile 
at one player  
may be classified as a suboptimal one at some other players.  
Increasing the length of exploration phase can reduce this uncertainty  
but it may induce a larger regret.  
\textit{How to determine the length of exploration phase?  
How to design an algorithm such that 
players can commit to an optimal arm pulling profile?}
We address them and our contributions are: 

\begin{itemize}
\item
In the offline setting where request distribution and 
reward distribution are given, 
we design a greedy algorithm which can locate one of the optimal 
arm pulling profiles with a computational complexity of $O(KM)$.  
We also design an iterative distributed algorithm for players 
to commit to a unique optimal arm pulling profile 
with a constant number of rounds in expectation.   
When there are multiple optimal arm pulling profiles, 
by using the same deterministic tie breaking rule in the greedy algorithm, 
all players can locate the same optimal arm pulling profile and 
then our committing algorithm can be applied.  

\item
In the online setting with unknown request distribution and 
reward distribution, we design an exploration strategy with a length 
such that each player can estimate one of the optimal arm pulling profiles  
with high probability.   
These estimates of the optimal arm pulling profile can be different across different players.  
We design an iterative distributed algorithm, 
which guarantees that players reach a consensus on the 
optimal arm pulling profile with only $M$ rounds.  
Players can then run the commit algorithm developed in the offline setting 
to commit to this consensus.     
Putting them together, we obtain an algorithm with a logarithmic regret.  
We conduct experiments to validate the efficiency of our algorithms.    
\end{itemize}

\section{Related Work}
\noindent
{\bf Stochastic multi-player MAB with collision. }  
The literature on multi-player MAB starts from a static 
(i.e., the number of players is fixed)
and informed collision (in each round, players know whether they encounter a collision or not) setting.  
In this setting, Liu \textit{et al.} \cite{Liu2010} proposed a time-division fair sharing algorithm, 
which attains a logarithmic total regret in the asymptotic sense.  
Anandkumar \textit{et al.} \cite{Anandkumar2011} proposed an algorithm 
with a logarithmic total regret in the finite number of rounds sense. 
Rosenski \textit{et al.} \cite{Rosenski2016}  
proposed a communication-free algorithm with constant regret in the high probability sense.  
Besson \textit{et al.} \cite{Besson2018} improved the regret lower bound,  
and proposed \emph{RandTopM} and \emph{MCTopM} which outperform existing algorithms empirically.     
The regret of these algorithms depends on the gaps of reward means.  
Lugosi \textit{et al.} \cite{Lugosi2018} suggested the idea of 
using collision information as a way to communicate and 
they gave the first square-root regret bounds that do not depend on the gaps of reward means. 
Boursier \textit{et al.} \cite{Boursier2019} further explored the idea of 
using collision information as a way to communicate 
and they proposed the SIC-MMAB algorithm, 
which attains the same performance as a centralized one.  
Inspired SIC-MMAB algorithm, 
Shi \textit{et al.} \cite{Shi2020} proposed 
the error correction synchronization involving communication algorithm, 
which attains the regret of a centralized one.  
Hanawal \textit{et al.} \cite{Hanawal2021} proposed the leader-follower framework 
and they developed a trekking approach, which attains a constant regret.  
Inspired by the leader-follower framework, 
Wang \textit{et al.} \cite{Wang2020} proposed the 
DPE1 (Decentralized Parsimonious Exploration) algorithm 
which attains the same asymptotic regret as that obtained by an optimal centralized algorithm.   
A number of algorithms were proposed for the static but unknown collision setting 
(in each round, players do not know whether they encounter a collision or not).  
In particular, Besson \textit{et al.} \cite{Besson2018} proposed a heuristic with nice empirical performance.  
Lugosi \textit{et al.} \cite{Lugosi2018} developed the first algorithm with theoretical guarantees, 
i.e., logarithmic regret, 
and an algorithm with a square-root regret type that does not depend on the gaps of the reward means.   
Shi \textit{et al.} \cite{Shi2020} identified a connection between  
communication phase without collision information 
and Z-channel model in information theory.  
They utilized this connection to design an algorithm 
with nice empirical performance over both 
synthetic and real-world datasets.  
Bubeck \textit{et al.} \cite{Bubeck2020} 
proposed an algorithm with near-optimal regret $O (\sqrt{T\log (T)}) $.  
They argued that the logarithmic term $\sqrt {\log (T)} $ is necessary.  
A number of algorithms were proposed for the dynamic (i.e., players can join or leave) 
and informed collision setting.  
In particular, Avner \textit{et al.} \cite{Avner2014} proposed an algorithm with a regret of $O(T^{2/3})$.  
Rosenski \textit{et al.} \cite{Rosenski2016} proposed the first 
 communication-free algorithms which attains a regret of $O(\sqrt{T})$.  
Boursier \cite{Boursier2019} proposed a SYN-MMAB algorithm 
with the logarithmic growth of the regret.   
Hanawal \textit{et al.} \cite{Hanawal2021} proposed an 
algorithm based on a trekking approach.  
The proposed algorithm attains a sub-linear regret with high probability.  
All the above works considered a homogeneous setting, i.e., 
all players have the same reward mean over the same arm.  
A number of works studied the heterogeneous setting, i.e., 
different players may have different reward mean over the same arm.  
Bistritz \textit{et al.} \cite{Bistritz2018} proposed the first algorithm which attains 
a total regret of order $O(\ln^2T)$.  
Magesh \textit{et al.} \cite{Magesh2019} proposed an algorithm which attains a 
total regret of $O(\log T)$.  
Mehrabian \textit{et al.} \cite{Mehrabian2020} proposed an algorithm 
which attains a total regret of $O(\ln(T))$ 
and it solved an open question raised in \cite{Bistritz2018}.  

\noindent
{\bf Stochastic multi-player MAB without collision. }  
The typical setting is that when collision happens, 
each player in the collision obtain an independent reward. 
Players can share their reward information using a communication graph.   
In this setting, Landgren \textit{et al.} \cite{Landgren2016} 
proposed a decentralized algorithm which utilizes a running consensus algorithm 
for agents to share reward information.  
Mart\'inez-Rubio \textit{et al.} \cite{Martinez2019} 
proposed a DD-UCB algorithm 
which utilizes a consensus procedure to estimate reward mean.  
Wang \textit{et al.} \cite{Wang2020} proposed DPE2 algorithm 
which is optimal in the symptotic sense and it 
outperforms DD- UCB \cite{Martinez2019}.    
Dubey \textit{et al.} \cite{Dubey2020}  
proposed MP-UCB to handle heavy tail reward.   

\noindent
{\bf Summary of difference.} 
Different from the above works, 
we formulate a new variant of multi-player 
multi-armed bandit (MAB) model 
to address the distributed selection problems.   
From a modeling perspective, 
our model captures stochastic arrival of request and 
request allocation policy of these applications.  
From an algorithmic perspective, 
our proposed algorithms presents new ideas in 
searching the optimal pulling profile, 
committing to optimal pulling profile, 
achieving consensus when different players 
have different estimates on the optimal 
arm pulling profile, etc.  

\section{Platform Model and Problem Formulation}
\label{gen_inst}
\subsection{The platform model}
We consider a platform composed of requests, players and 
a platform operator.  
We use a discrete time system indexed by $t \in \{1, \ldots, T\}$, 
where $T \in \mathbb{N}_+$, to model this platform.  
The arrival of requests is modeled by a finite set of arms denoted by 
$\mathcal{M} \triangleq \{1,\ldots, M\}$, 
where $M \in \mathbb{N}_+$.  
Each arm can be mapped as a pickup location of ride sharing applications  
or a pickup port of food delivery applications.   
Each arm $m \in \mathcal{M}$ 
is characterized by a pair of random vectors $(\bm{D}_m, \bm{R}_m)$, where 
$
\bm{D}_m 
\triangleq 
[D_{t,m} : t = 1, \ldots, T]
$ 
and 
$
\bm{R}_m 
\triangleq 
[R_{t,m} : t = 1, \ldots, T]
$
model the stochastic request and reward of arm $m$ across time slots 
$1, \ldots, T$.   
More concretely, the random variable $D_{t,m}$ models 
the number of requests arrived at arm $m$ in time slot $t$, 
and the support of $D_{t,m}$ is 
$\mathcal{D} \triangleq \{1, \ldots, d_{\max}\}$, 
where $d_{\max} \in \mathbb{N}_+$.  
Each request can be mapped as a ride sharing request 
or a food delivering request.  
We consider a stationary arrival of requests, 
i.e., $D_{1,m}, \ldots, D_{t,m}$ are independent and identically distributed (IID)
random variables.  
Note that in each time slot unserved requests will be dropped.  
This captures the property of ride sharing like applications that 
a customer may not wait until he is served, 
but instead he will cancel the ride sharing request and try other 
alternatives such as buses if he is not severed in 
a time slot.  
Let 
$\bm{p}_m 
\triangleq 
[p_{m,d}: \forall d \in \mathcal{D}]$ denote the probability mass function (pmf) 
of these IID random variables $D_{1,m}, \ldots, D_{t,m}$, formally
$  
p_{m,d} 
= 
\mathbb{P} [D_{t,m} = d],  
\forall d \in \mathcal{D}, 
m \in \mathcal{M}.  
$
In time slot $t$, the rewards of $D_{t,m}$ requests are 
IID samples of the random variable $R_{t, m}$.  
Without loss of generality we assume the support of $R_{t,m}$ is $[0,1]$.  
The rewards $R_{1,m}, \ldots, R_{t,m}$ are IID random variables.   
We denote the mean of these IID random variables $R_{1,m}, \ldots, R_{t,m}$ as 
$
\mu_m = \mathbb{E} [R_{t,m}],  
\forall m \in \mathcal{M}.   
$
For the ease of presentation, 
we denote the reward mean vector as 
$
\bm{\mu} 
\triangleq 
[\mu_m : \forall m \in \mathcal{M}] 
$
and probability mass matrix as 
$
\bm{P} 
\triangleq 
[\bm{p}_1, \ldots, \bm{p}_M]^{\text{T}}.  
$
Both $\bm{P}$ and $\bm{\mu}$ are unknown to players.   
 
We consider a finite set of players denoted by 
$\mathcal{K} \triangleq \{1, \ldots, K\}$, 
where $K \in \mathbb{N}_+$.  
Each player can be mapped as a driver in ride sharing applications, 
or a deliverer in food deliverer applications.  
In each time slot $t$, each player is allowed to pull only one arm.  
Let $a_{t,k} \in \mathcal{M}$ denote the action 
(i.e., the arm pulled by player $k$) of player $k$ in time slot $t$.  
We consider a distributed setting that 
players can not communicated with each other.  
Let 
$
n_{t,m}
\triangleq 
\sum_{ k \in \mathcal{K} }  
\mathbbm{1}_{ \{ a_{t,k} = m \} }
$ 
denote the number of players who pull arm $m$.  
The $n_{t,m}$ satisfies that $\sum_{m \in \mathcal{M}} n_{t,m} = K$. 
Namely, all players are assigned to arms.   
Recall that in round $t$, the number of requests arrived at arm $m$ is $D_{t,m}$.   
These $D_{t,m}$ requests will be allocated to $n_{t,m}$ players 
randomly (our algorithm can be applied to other assignment policies also).   
Regardless of the details of the allocation policy, 
two desired properties of the allocation is: 
(1) if the number of requests is larger than 
the number of players, i.e., $D_{t,m} \geq n_{t,m}$, 
then each player serves one request and $(D_{t,m} - n_{t,m})$ 
request remains unserved, 
(2) if the number of requests is smaller than 
the number of players, i.e., $D_{t,m} \leq n_{t,m}$, 
then only $D_{t,m}$ players can serve requests 
(one player per request) and $(n_{t,m} - D_{t,m})$ remains idle.

Let $\bm{n}_t \triangleq [n_{t,m} : \forall m \in \mathcal{M}]$ 
denote the action profile in time slot $t$.  
Let $\widetilde{\bm{D}}_t \triangleq [D_{t,m} : \forall m \in \mathcal{M}]$ denote
the request arrival profile in time slot $t$.  
At the end the each time slot, the platform operator 
makes $\bm{n}_t$ and $\widetilde{\bm{D}}_t$ public to all players.  
The platform operator ensures that 
each player is allowed to serve at most one request.  
When the number of requests exceeds 
the number of players who pull the arm, 
each player serves one request and the remaining requests are unserved, 
and on the contrary,  
the remaining players will be idle. 

\subsection{Online learning problem}
Let $U_m (n_{t,m} , \bm{p}_m, \mu_m)$ denote 
the total reward 
earned by $n_{t,m}$ players pull arm $m$.   
Then,  it can be expressed as: 
$
U_m (n_{t,m} , \bm{p}_m, \mu_m)
= 
\mu_m 
\mathbb{E} 
\left[ 
\min{\{n_{t,m},D_{t,m}\}}
\right]
$.  
Denote the total reward 
for earned by all players in time slot $t$ as
$
U \left(\bm{n}_t,\bm{P}, \bm{\mu}\right) 
\triangleq 
\sum\nolimits_{m\in\mathcal{M}} U_m (n_{t,m} , \bm{p}_m, \mu_m).      
$ 
The objective of players is to maximize the total reward 
across $T$ time slots, i.e., 
$\sum^T_{t=1} U \left(\bm{n}_t,\bm{P}, \bm{\mu}\right)$.  
The optimal arm pulling profile of players is  
$
\bm{n}^{\ast} 
\in 
\arg\max_{ \bm{n} \in \mathcal{A} } 
U \left(\bm{n}, \bm{P}, \bm{\mu}\right),
$
where 
$
\mathcal{A} 
\triangleq 
\{
(n_1, \ldots, n_M) 
\big| 
n_m \in \mathcal{K} \cup \{0\}, 
\sum_{m \in \mathcal{M}} n_m = K
\}   
$
is defined as a set of all arm pulling profiles.  
There are $|\mathcal{A}| = {M+K-1 \choose M-1}$ possible arm pulling profiles, 
which poses a computational challenge in searching  
the optimal arm pulling profile.    
Furthermore, both the probability mass matrix $\bm{P}$ 
and the reward mean vector $\bm{\mu}$ are unknown to players 
and players can not communicate to each other.  
Denote 
$
\mathcal{H}_{t,k}
\triangleq 
(a_{1,k}, X_{1,k}, \bm{n}_1, \widetilde{\bm{D}}_1, 
\ldots, 
a_{t,k}, X_{t,k}, \bm{n}_t, \widetilde{\bm{D}}_t)  
$ 
as the historical data available to player $k$ up to time slot $t$.   
Each player has access to his own action history and 
reward history as well as the arm pulling profile history and 
request arrival profile which are made public by the platform operator.  
Denote the regret as 
$
R_T
\triangleq
\sum\nolimits_{t=1}^T 
( 
U\left(\bm{n}^\ast, \bm{P}, \bm{\mu}\right)
-
\mathbb{E}\left[U\right(\bm{n}_t,\bm{P},\bm{\mu}\left)\right]
).  
$

\section{The Offline Optimization Problem}
\subsection{Searching the optimal arm pulling profile} 
We define the marginal reward gain function as: 
$
\Delta_m (n) 
\triangleq 
U_m (n+1 , \bm{p}_m, \mu_m) 
-
U_m (n , \bm{p}_m, \mu_m).  
$

\begin{lemma}
The $\Delta_m (n) = \mu_m   P_{m,n+1}$, 
where $P_{m,n+1} \triangleq \sum^{d_{\max}}_{ d = n+1} p_{m, d}$.
Furthermore,  
$\Delta_m (n+1) \leq \Delta_m (n)$.  
\label{lem:offline:maringalgain}
\end{lemma}  

Based on Lemma \ref{lem:offline:maringalgain}, 
Algorithm \ref{algo:greedy} searches the optimal arm pulling profile 
by sequentially adding $K$ players one by one to pull arms.  
More specifically, players are added to arms sequentially according to 
their index in ascending order.  
Player with index $k$, is added to the arm with the 
largest marginal reward gain given 
the assignment of players indexed by $1, \ldots, k-1$.  
Whenever there is a tie, breaking it by an arbitrary tie breaking rule.
When all players are added to arms, 
the resulting arm pulling profile is returned as an output.  
For simplicity, 
denote $\widetilde{\bm{P}}=[\widetilde{\bm{P}}_1,\ldots,\widetilde{\bm{P}}_M]^T$, 
where $\widetilde{\bm{P}}_m = [P_{m,k}:\forall k \in \mathcal{K}]$.  

\begin{algorithm}[h]  
    \caption{\texttt{OptArmPulProfile} $(\bm{\mu}, \widetilde{\bm{P}})$}
    \label{algo:greedy}
    \begin{algorithmic}[1] 
    \STATE $\Delta_m \leftarrow \mu_m, \forall m \in \mathcal{M}$, 
    \hspace{0.08 in}
    $n_{\text{greedy},m} \leftarrow 0, \forall m \in \mathcal{M}$
    \FOR{$k=1, \ldots, K$}
    \STATE $i = \arg\max_{m \in \mathcal{M}} \Delta_m $ (if there is a tie, breaking it by an arbitrary tie breaking rule)
    \STATE $n_{\text{greedy},i}  \leftarrow   n_{\text{greedy},i} + 1$, 
    \hspace{0.08 in}
    $\Delta_i \leftarrow \mu_i P_{i,n_{\text{greedy},i} }$
    \ENDFOR 
    \RETURN $\bm{n}_{\text{greedy}} = [n_{\text{greedy}, m} : \forall m \in \mathcal{M}]$
      \end{algorithmic}  
\end{algorithm} 
    
\begin{theorem} 
\label{thm:offline:GreedyGuarantee}
The output $\bm{n}_{\text{greedy}}$ of Algorithm \ref{algo:greedy} satisfies that 

$
\bm{n}_{\text{greedy}} 
\in 
\arg\max_{ \bm{n} \in \mathcal{A} } 
U \left(\bm{n}, \bm{P}, \bm{\mu}\right).  
$
The computational complexity of $\bm{n}_{\text{greedy}}$ of 
Algorithm \ref{algo:greedy} is $O(K M)$.  
\end{theorem}
\noindent

\noindent
Theorem \ref{thm:offline:GreedyGuarantee} states that 
Algorithm \ref{algo:greedy} locates one of the optimal 
arm pulling profiles.  
Compared to the computational complexity of exhaustive search  
${M+K-1 \choose M-1}$, 
one can observe that Algorithm \ref{algo:greedy} drastically reduces the 
computational complexity.  
Furthermore, the computational complexity of 
Algorithm \ref{algo:greedy} scales linearly with respect to $K$, 
and scales in linearly with respect to M.  
Namely, Algo. \ref{algo:greedy} is applicable to a large 
number of players or arms.

\subsection{Committing to optimal arm pulling profile}
Recall that players can not communicate with each other in pulling arms.  
Now, we design algorithms such that players commit to 
the optimal arm pulling profile without communication with each other.  

We first consider the case that the number of optimal arm pulling profiles 
is unique.  
We will generalize to handle multiple optimal pulling profiles later.  
Note that the probability mass matrix $\bm{P}$ and the mean vector 
$\bm{\mu}$ is known to players.  
Each player first applies Algorithm \ref{algo:greedy} 
to locate the unique optimal pulling profile $\bm{n}^\ast$.  
In each round $t$, player $k$ selects arm based on 
$\bm{n}^\ast$ and $\bm{n}_1, \ldots, \bm{n}_{t-1}$.  
Our objective is that $n^\ast_m$ players commit to arm $m$.  

Let $c_{t, k} \in \{0\} \cup \mathcal{M}$ 
denote the index of the arm that player $k$ commits to.  
The $c_{t, k}$ is calculated from $\bm{n}_1, \ldots, \bm{n}_{t}$  
and $\bm{n}^\ast$ as follows.  
Initially, player $k$ sets $c_{0, k} = 0$ representing that he has not 
committed to any arm yet.  
In each time slot $t$, after $\bm{n}_{t}$ is published, 
each player $k$ calculates $c_{t, k}$ based on $\bm{n}_t$, 
$\bm{n}^\ast$ and $c_{t-1, k}$ as follows: 
\begin{equation}
	c_{t, k} 
	= 
	\mathbbm{1}_{ \{ c_{t-1, k} \neq 0 \} } 
	c_{t-1, k} 
	+ 
	\mathbbm{1}_{ \{ c_{t-1, k} = 0 \} } 
	\mathbbm{1}_{ \{ n_{t,a_{t,k}} \leq n^\ast_{a_{t,k}} \} } 
	a_{t,k}.   
	\label{eq:UpdataCommit}
\end{equation}
Equation (\ref{eq:UpdataCommit}) states that if 
player $k$ has committed to an arm, i.e., $c_{t-1, k} \neq 0$, 
this player will stay committed to this arm.  
In other words, once a player commits to an arm, 
he will keep pulling it in all remaining time slots.  
If player $k$ has not committed to any arm yet, 
i.e., $c_{t-1, k} = 0$, 
player $k$ commits to the arm he pulls $a_{t,k}$,  
only if the number of players pull the same arm  
does not exceed the number of players required by this arm, 
i.e., $n_{t,a_{t,k}} \leq n^\ast_{a_{t,k}}$.  
In a nutshell, in each round only the players who have not committed to any arm 
need to selecting different arms.  

To assist players who have not committed to any arm selecting arms, 
for each arm, each player keeps a track of the number of players 
who have committed to it.   
Let $n^+_{t,m}$ denote the number of players committing to arm $m$ up to 
time slot $t$.  
Initially, no players commit to each arm, 
i.e., $n^+_{0,m} = 0, \forall m \in \mathcal{M}$.  
After round $t$, each player uses the following rule to update $n^+_{t,m}$:  
\begin{equation}
	n^+_{t,m} 
	= 
	\mathbbm{1}_{ \{ n_{t,m} \leq n^\ast_m \} } 
	n_{t,m} 
	+ 
	\mathbbm{1}_{ \{ n_{t,m} > n^\ast_m \} } 
	n^+_{t-1,m}.   
	\label{eq:UpdataCommitNum}
\end{equation}
Equation (\ref{eq:UpdataCommitNum}) states an update rule that 
is consistent with Equation (\ref{eq:UpdataCommit}). 
More concretely, if the number of players $n_{t,m}$ 
pull arm $m$ does not exceed  
the optimal number of players $n^\ast_m$ for arm $m$, 
then all these $n_{t,m}$ players commit to arm $m$.  
Otherwise, as $n_{t,m} > n^\ast_m$, it is difficult for players to decide 
who needs to commit to arm $m$ without communication.  
According to Equation (\ref{eq:UpdataCommit}), 
the commitment status of players pulling arm $m$ does not update, 
resulting that the number of player commit to arm $m$ remains unchanged.  
Equation (\ref{eq:UpdataCommitNum}) implies that $n^+_{t,m} \leq n^\ast_m$.  

In time slot $t$, each player $k$ selects arm based on $c_{t-1,k}$ 
and $\bm{n}^+_{t-1} 
\triangleq 
(n^+_{t-1, m} : m \in \mathcal{M})
$  
as follows.  
If player $k$ has committed to an arm, i.e, $c_{t-1,k} \neq 0$, 
this player sticks to the arm that he commits to.  
Otherwise, player $k$ has not committed to any arm yet, 
and he needs to select an arm.  
To achieve this, player $k$ first calculates the number of players 
that each arm $m$ lacks, 
which is denoted by $n^-_{t-1, m} \triangleq n^\ast_m - n^+_{t-1, m}$.  
Note that $n^\ast_m - n^+_{t-1, m} \geq 0$ 
because Equation (\ref{eq:UpdataCommitNum}) implies 
that $n^+_{t-1,m} \leq n^\ast_m$.  
Then, player $k$ selects an arm with a probability proportional 
to the number of players that the arm lacks, 
i.e., selects arm $m$ with probability 
$
n^-_{t-1, m}
/ 
N^-_{t-1}
$, 
where $N^-_{t-1} {\triangleq} \sum_{m \in \mathcal{M}} n^-_{t-1, m}$.  
We summarize the arm selection strategy as follows:  
$
\mathbb{P} 
[
a_{t,k} 
= 
m
]
=
\mathbbm{ 1 }_{ 
	\{ c_{t-1,k} \neq 0 \} 
}
\mathbbm{ 1 }_{ 
	\{ c_{t-1,k} = m \} 
}
+
\mathbbm{ 1 }_{ 
	\{ c_{t-1,k} = 0 \} 
}
n^-_{t-1, m}
/ 
N^-_{t-1}.  
$
Players use the same committing strategy, 
Algorithm \ref{algo:commitOpt} uses player $k$ as an example 
to outline the above committing strategy, 
where $T_{\text{start}}$ denotes the 
index of the time slot that players start committing.  

\begin{algorithm}[h]  
	\caption{ {\small \texttt{CommitOptArmPulProfile} $(k,\bm{n}^\ast, T_{\text{start}})$ }}
	\label{algo:commitOpt}
	\begin{algorithmic}[1] 
		\STATE $c_k {\leftarrow} 0, {\forall} k {\in} \mathcal{K}$, 
		$n_{t,m}^- {\leftarrow} n_m^{\ast}, \forall m {\in} \mathcal{M}$, 
		$n_{t,m}^+ {\leftarrow} 0, \forall m {\in} \mathcal{M}$
		
		\FOR{$t=T_{\text{start}}, \ldots, T$}
		\IF{$ c_k \neq 0$ } 
		\STATE $a_{t+1,k} \leftarrow a_{t,k}$
		
		\ELSIF{$n_{a_{t,k}}^\ast  \geq n_{t,a_{t,k}}$}
		\STATE $c_k \leftarrow a_{t,k}$, 
		\hspace{0.08in}
		$a_{t+1,k} \leftarrow a_{t,k}$
		
		\ELSE
		\FOR{$m=1, \ldots, M$}
		\IF{ $n_m^\ast\geq n_{t,m}$}
		\STATE {$n_{t,m}^+ \leftarrow n_{t,m}$}, 
		\hspace{0.08in} 
		{$n_{t,m}^- \leftarrow n_m^\ast - n_{t,m}^+$}
		\ENDIF
		\ENDFOR
		\STATE{$N_t^- = \sum_{m\in\mathcal{M}} n_{t,m}^-$}.    
		{$a_{t+1,k} {\leftarrow} m$, w.p. $n_{t,m}^- / N_t^-$}
		\ENDIF
		\STATE player $k$ pulls arm $a_{t+1,k}$, 
		\hspace{0.08in}
		player $k$ receive $\bm{n}_{t+1}$
		\ENDFOR
	\end{algorithmic}  
\end{algorithm}


\begin{theorem}
	In expectation, Algorithm \ref{algo:commitOpt} terminates in a number of rounds bounded by $T_{\text{commit}}$, 
	where
	$
	T_{commit}
	\triangleq
	\sum_{m\in\mathcal{M}}
	{1} / 
	\left[
	{ 
		{
			K 
			\choose
			n^\ast_{m}
		}
		\left(
		\frac{n^\ast_{m}}{K}
		\right)^{ n^\ast_{m} } 
		\left(
		1
		-
		\frac{n^\ast_{m}}{K}
		\right)^{K- n^\ast_{m} }
	}
	\right]
	$.  
	\label{thm:offline:ConvergenceAlgoComit}
\end{theorem}

\noindent
Theorem \ref{thm:offline:ConvergenceAlgoComit} states that 
Algorithm \ref{algo:commitOpt} terminates in a constant number 
of rounds in expectation.  

Now we consider the case that there are multiple optimal arm pulling profiles.  
Each player can use the same deterministic 
tie breaking rule in Algorithm \ref{algo:greedy} to locate the same optimal 
arm pulling profile.  
For example, whenever there is a tie among arms, 
select the arm with the smallest index.  
Through this, players has the same optimal arm pulling profile at hand.  
Then, players can follows Algorithm \ref{algo:commitOpt}  
to commit to the optimal arm pulling profile.  

\section{Online Learning Algorithm}
\label{headings}
Similar with previous works \cite{Bistritz2018,Rosenski2016}, 
we address the online setting with unknown 
$\bm{\mu}$ and $\bm{P}$ 
via the ETC framework.   

\noindent
\subsection{Exploration Phase}   
In the exploration, each player aims to estimate the probability  
$\widetilde{\bm{P}}$ and 
the mean vector $\bm{\mu}$.  
We consider a random exploring strategy that 
in each decision round, 
each player randomly selects an arm.   
Let $T_0$ be the length of the exploration phase.  
Each player uses the same exploration strategy, 
and Algorithm \ref{algo:Explore} uses player $k \in \mathcal{K}$ 
as an example to illustrate our exploration strategy.     
\begin{algorithm}[h]  
    \caption{\texttt{Explore}$(k,T_0)$}
    \label{algo:Explore}
    \begin{algorithmic}[1] 
    \STATE $S_m \leftarrow 0$, 
    $\widehat{P}_{m,d}^{(k)} \leftarrow 0$, 
    $\widetilde{c}_m \leftarrow 0, \forall m \in \mathcal{M}$
    \FOR{$t=1, \ldots, T_0$ }
    \STATE $a_{t,k} \sim U(1, \ldots, M)$.   
    Player $k$ receives $\widetilde{\bm{D}}_t$ and $X_{t,k}$ 
    \IF{$X_{t,k} \neq \text{null}$}
    \STATE $S_{a_{t,k}} \leftarrow S_{a_{t,k}}+  X_{t,k}$, 
    \hspace{0.08in}
    $\widetilde{c}_{a_{t,k}} \leftarrow \widetilde{c}_{a_{t,k}}+ 1 $
    \ENDIF 
    \STATE 
    $\widehat{P}_{m,d}^{(k)}\leftarrow \widehat{P}_{m,d}^{(k)} 
    + 
    \mathbbm{1}_{ \{\widetilde{D}_{t,m}\geq d \}  }, 
    \forall d \in \mathcal{D}, 
    m \in \mathcal{M}
    $ 
    \ENDFOR 
    \STATE {$\widehat{\mu}_m^{(k)} \leftarrow S_m / \widetilde{c}_m, \forall m \in \mathcal{M}$}, 
    \hspace{0.08in}
    {$\widehat{P}_{m,d}^{(k)} \leftarrow \widehat{P}_{m,d}^{(k)} / T_0, 
    \forall d, m$}
    \RETURN {$\widehat{\bm{\mu}}^{(k)} = [\widehat{\mu}_m^{(k)}:  \forall m ]$,
     $\widehat{\bm{P}}^{(k)} = [\widehat{P}_{m,d}^{(k)}: \forall m, d]$}
    \end{algorithmic}  
\end{algorithm}

\begin{lemma}
	The output of 
	Algorithm \ref{algo:Explore} satisfies: 
	\begin{align*} 
		\mathbb{P} 
		\Big[
		&
		\forall k \in \mathcal{K}, 
		d \in \mathcal{D}, 
		m \in \mathcal{M}, \\
		&\widehat{P}^{(k)}_{m,d}
		- 
		P_{m,d}
		\leq 
		\sqrt{
			( \ln \delta_1^{-1} ) / ( 2 T_0 )
		}, \\
		&\widehat{\mu}^{(k)}_m - \mu_m
		\leq
		\sqrt{  ( \ln{\delta_2^{-1}} ) / ( c_m T_0 )   }
		\Big]
		\\
		& 
		\geq 
		1 - MK\delta_1 - MK\delta_2 
		- K\sum_{m\in \mathcal{M}}
		\exp\left(
		- T_0 c_m / 8 
		\right), 
	\end{align*} 
	where 
	$
	c_m 
	\triangleq 
	\frac{1}{M} 
	\sum_{d \in \mathcal{D}} 
	p_{m,d} 
	\sum^{K-1}_{n =0}  
	{K-1 \choose n} M^{- (K-1)}
	\min \left\{ 1, \frac{d}{n} \right\}  
	$
	denotes the probability that 
	a player gets one reward from arm $m$.  
	\label{lem:Online:exploration}
\end{lemma}

Lemma \ref{lem:Online:exploration} quantifies the impact of the length of exploration phase 
on the accuracy of estimating $\bm{\mu}$ and $\bm{P}$.  
Each player uses the plugging method to estimate the 
optimal arm pulling profile 
and we denote the estimate by 
$\widehat{\bm{n}}^\ast (k)
= \texttt{OptArmPulProfile} (k, \widehat{\bm{\mu}}^{(k)}, \widehat{\bm{P}}^{(k)})
$.  
We need the following notations to characterize $\widehat{\bm{n}}^\ast (k)$.  
Let $\Delta^{(1)},\ldots, \Delta^{(KM)}$ denote a ranking list of 
$\Delta_m (n), \forall m = 1, \ldots, M, n = 0, \ldots, K-1$ 
such that $\Delta^{(i)} \geq \Delta^{(i+1)}$.    
Denote $
\ell^- = 
|
\{
j | j > K, \Delta^{(j)} =\Delta^{(K)}
\}
|
$, 
$
\ell^+ = 
|
\{
j | j < K, \Delta^{(j)} =\Delta^{(K)}
\}
|
$
and $
\gamma 
\triangleq 
\min{
\{ 
\Delta^{(K)} 
-
\Delta^{(K+\ell^-+1)}, 
\Delta^{(K - \ell^+ - 1)} - \Delta^{(K)}
\}}  
$.

\begin{theorem} 
	By setting $T_0$ to 
	\begin{align*}
		&T_0\\ 
		&{=} 
			\max_{m\in \mathcal{M}}(
			(2/c_m)
			\ln{\delta_1^{-1}}
			\ln{\delta_2^{-1}}
		/
			\\&
			(
			\sqrt{
				0.5\ln{\delta_1^{-1}}
				+{c_m}^{-1}
				\ln{\delta_2^{-1}}
				+(2+2\gamma)
				(2c_m)^{-1/2}
				\sqrt{\ln{\delta_1^{-1}}\ln{\delta_2^{-1}}}
			}\\
			&- \sqrt{0.5
				\ln{\delta_1^{-1}}}
			- \sqrt{c_m^{-1}
				\ln{\delta_2^{-1}}}
			)^2)
		, 
	\end{align*}
	we have 
	\begin{align*}
		&\mathbb{P}
		[
		\forall k {\in} \mathcal{K}, 
		\widehat{\bm{n}}^\ast (k) 
		{\in} 
		\arg\max_{ \bm{n} \in \mathcal{A} } 
		U \left(\bm{n}, \bm{P}, \bm{\mu}\right) 
		]\\
		&{\geq} 
		1 {-} MK\delta_1 {-} MK\delta_2
		{-} K\sum_{m\in \mathcal{M}}\exp\left(
		- 
		T_0c_m / 8
		\right).
	\end{align*}
	\label{thm:estOptProfile}
\end{theorem}

\subsection{Committing Phase} 
When the optimal arm pulling profile is not unique, 
it is highly likely that players have different estimates 
on the optimal arm pulling profile, 
i.e., $\exists k, \tilde{k}$ such that 
$\widehat{\bm{n}}^\ast (k) \neq \widehat{\bm{n}}^\ast (\tilde{k})$.  
This creates a challenge in committing to the optimal arm pulling profile.  
To address this challenge, we make the following observations.  
An element $\Delta_m (n)$ is a borderline element 
if $\Delta_m (n) = \Delta^{(K)}$.   

\begin{lemma} 
    Suppose $p_{m,d} > 0$ holds for all $m \in \mathcal{M}$ 
    and $d \in \mathcal{D}$.  
    Suppose $\bm{n}^\ast$ and $\widetilde{\bm{n}}^\ast$ denote to 
    optimal arm pulling profiles.  
    Then,  
    $
    |n^\ast_m - \widetilde{n}^\ast_m| \leq 1, 
    \forall m \in \mathcal{M} 
    $
    and if $n^\ast_m \neq \widetilde{n}^\ast_m$, 
    $\Delta_m ( \max\{n^\ast_m, \widetilde{n}^\ast_m\} )$ 
    is a borderline element.  
    \label{lem:disagree}
\end{lemma}
Lemma \ref{lem:disagree} implies that two optimal arm pulling profiles only 
possible to disagree on the borderline elements.  
Algorithm \ref{algo:Consistency} uses player $k$ as an example to illustrate 
our consensus algorithm, which enables players to reach a consensus on 
the optimal arm pulling profile.     
Note that Algorithm \ref{algo:Consistency} focuses on the case 
that each player has an accurate estimate of 
the optimal arm pulling profile, 
i.e., $\widehat{\bm{n}}^\ast (k) 
\in 
\arg\max_{ \bm{n} \in \mathcal{A} } 
U \left(\bm{n}, \bm{P}, \bm{\mu}\right)$, 
but their estimates can be different.  
First, all players run $M$ rounds 
to identify disagreements in their estimates of optimal arm pulling profiles 
(step 2 to 11).  
In each round, they check disagreements on one arm, 
if they identify one disagreement, each player records 
the corresponding borderline elements.  
After this phase, each player eliminates the 
identified borderline elements from its estimate on the optimal arm pulling profile 
(step 12 to 16).  
After this elimination, players agree on their remaining arm pulling profile, but 
this arm pulling profile only involves a number of players less than $K$.  
Thus, finally, each player adding the same number of 
borderline elements as the number of eliminated ones back, 
using the same rule, 
i.e., guaranteeing that all players 
add the same borderline elements back 
(step 17 and 18).  

\begin{algorithm}[h]  
    \caption{\texttt{Consensus}$(k, \widehat{\bm{n}}^\ast (k))$}  
    \label{algo:Consistency}
    \begin{algorithmic}[1]  
    \STATE $\bm{v}_{\text{board},k} \leftarrow \bm{0}$, 
    $\mathcal{V}_{\text{board},k} \leftarrow \emptyset$, 
    Num$\leftarrow$0
    \FOR{$t=T_0+1, \ldots, T_0+M$ }
    \STATE player $k$ pulls arm $(\widehat{n}^\ast_{t -T_0}(k) \mod M +1 )$  
    
    \IF{$\max  \{ m | n_{t,m} > 0 \} - \min  \{ m | n_{t,m} > 0 \} == 1$ } 
    \STATE Update borderline elem. 
    $\bm{v}_{\text{board},k} {\leftarrow} 
    (\bm{v}_{\text{board},k}, t -T_0) $ 
    
    \STATE $\mathcal{V}_{\text{board},k} \leftarrow 
    \mathcal{V}_{\text{board},k}  \cup 
    \{ ( t -T_0, 
    \widehat{n}^\ast_{t-T_0}(k) 
    + 
    \mathbbm{1}_{
    \{
    \widehat{n}^\ast_{t -T_0}(k) \mod M +1 
    = 
    \min  \{ m | n_{t,m} > 0 \}
    \}
    } ) \}
    ) \}$
    \ELSIF{$\max  \{ m | n_{t,m} > 0 \} - \min  \{ m | n_{t,m} > 0 \} > 1$}
    \STATE 
    $\bm{v}_{\text{board},k} \leftarrow 
    (\bm{v}_{\text{board},k}, t -T_0), \forall k \in \mathcal{K}$ 
    
    \STATE $\mathcal{V}_{\text{board},k} \leftarrow 
    \mathcal{V}_{\text{board},k}  \cup \{ ( t-T_0, \widehat{n}^\ast (k) 
    + \mathbbm{1}_{
    \{
    \widehat{n}^\ast_{t-T_0}(k) \mod M +1 
    = 
    \max  \{ m | n_{t,m} > 0 \}
    \}
    } ) \}, 
    \forall k \in \mathcal{K}$
    \ENDIF
    \ENDFOR
    
    \FOR {$m = 1, \ldots, M$}
    
    \IF {$\{( m, \widehat{n}^\ast_m (k) )\} \in \mathcal{V}_{\text{board},n}$ } 
    \STATE $\widehat{n}^\ast_m (k) \leftarrow \widehat{n}^\ast_m(k) - 1$, 
    \hspace{0.18 in} 
    Num$\leftarrow$Num $+1$
    \ENDIF
    \ENDFOR
    \STATE $\bm{v}_{\text{board},n} \leftarrow$ $\bm{v}_{\text{board},n}$ sorted in descending order 
    
    \STATE $\widehat{n}^\ast_{ \bm{v}_{\text{board},k} (i) } (k) 
    \leftarrow \widehat{n}^\ast_{ \bm{v}_{\text{board},k} (i) } (k) + 1, \forall i = 1, \ldots, \text{Num}$
    \RETURN {$\widehat{\bm{n}}^\ast (k)=[\widehat{n}^\ast_m(k):\forall m\in\mathcal{M}]$}
    
      \end{algorithmic}  
\end{algorithm}

\begin{theorem}
    Suppose 
    $
    \widehat{\bm{n}}^\ast (k) 
    {\in} 
    \arg\max_{ \bm{n} \in \mathcal{A} } 
    U \left(\bm{n}, \bm{P}, \bm{\mu}\right), 
    \forall k \in \mathcal{K}. 
    $  
    and $M\geq 3$.  
    Algo. \ref{algo:Consistency} reaches a consensus, 
    i.e., each player has the same optimal arm pulling profile.   
    \label{thm:consensus}
\end{theorem}
    
Theorem \ref{thm:consensus} states that if 
each player has an optimal arm pulling profile, 
Algorithm \ref{algo:Consistency} guarantees that 
they reach a consensus on the optimal arm pulling profile.  
With this consensus, Algorithm \ref{algo:commitOpt} 
can be applied for players to commit to 
this consensus.  

\subsection{Putting Them Together and Regret Analysis} 
Puting all previous algorithms together, 
Algo. \ref{algo:sumup} outlines our algorithm 
for the online setting.   
The following theorem bounds the regret of Algo. \ref{algo:sumup}.  

\begin{algorithm}[h]  
    \caption{Distributed learning algorithm for multi-player MAB with stochastic requests}
    \label{algo:sumup}
    \begin{algorithmic}[1] 
    \STATE 
    Exploration:   
    $( \widehat{\bm{\mu}}^{(k)}, \widehat{\bm{P}}^{(k)} ) 
    \leftarrow$ \texttt{Explore}$(k,T_0)$
    \STATE 
    Estimate optimal arm pulling profile:  \\
    $ 
    \bm{n}_{\text{greedy}}(k) 
    \leftarrow$ \texttt{OptArmPulProfile}$(\widehat{\bm{\mu}}^{(k)}, \widehat{\bm{P}}^{(k)})$
    \STATE 
    Consensus: 
    $\widetilde{\bm{n}}^\ast   
    \leftarrow \texttt{Consensus} (k,\bm{n}_{\text{greedy}}(k))$
    \STATE 
    Committing to consensus: \\
    $\texttt{CommitOptArmPulProfile}(k,\widetilde{\bm{n}}^\ast, T_0 + M  )$
    \STATE 
    Sticking to the committed consensus
    \end{algorithmic}  
    \end{algorithm}

    \begin{theorem}
    The regret of Algorithm \ref{algo:sumup} is bounded by 
    \begin{align*}
    R_T
    &  \leq 
    T_0K+MK+KT_\text{commit} + \\
    & (T-T_0-M-T_\text{commit})K^2
    [M\delta_1+M\delta_2+
    \sum_{m\in \mathcal{M}}\exp\left(
    -T_0c_m / 8
    \right)
    ] 
    .
    \end{align*}  
    If  
    $
    T_0 {=}
    \max \!\!
    \big\{ 
    2c^{-1}_m \ln T
    /
    (
    \sqrt{
    0.5   {+}    c_m^{-1}   {+}    (2+2\gamma)  (2c_m)^{-1/2}
    } 
     {-}
    \sqrt{0.5}
    \\
    - c_m^{-1/2}
    )^{2}
    ,
    \max_{m\in \mathcal{M}}
    8c^{-1}_m\ln T
    \big\} 
    $, then
    $R_T \leq O(\ln T)$.  
    \label{thm:online:regret}
\end{theorem}

Theorem \ref{thm:online:regret} states the 
impact of $T_0$ on the regret.  
It states that by selecting appropriate length of exploration, 
Algorithm \ref{algo:sumup} can achieve a logarithmic regret.  

\section{Experiment}
\subsection{Experiment Setup}
{\bf Parameter setting}.  
Unless we vary them explicitly, we consider the following default parameters: $T  = 10^4$, $K=150$ players, $M=50$ arms, 
each arm's reward have same standard deviation $\sigma=0.1$ and $d_{\max} =50$. 
The mean reward of each arm is drawn from $[0,1]$ uniformly at random. 
We generate the reward of each request via a normal distribution.   
For each arm, 
we first generate $d_{\max}$ numbers from $[0,1]$ uniformly at random.  
Then we normalize these number such that their sum equals one.  
We use these normalized numbers as the probability mass 
of one arm.  
Repeating this process for all arms we obtain the 
probability mass matrix.   

\noindent
{\bf Baseline and metrics}.  
We consider the following two baselines.   
(1) {\bf MaxAvgReard}, 
where each player pulls arm with the largest average reward 
estimated from the collected historical rewards.   
(2) {\bf SofMaxReward}, 
where each player selects arm $m$ with probability 
proportional to the exponential of the average reward 
estimated from the collected historical rewards, 
i.e., softmax of average reward.  
We consider two metrics, 
i.e., regret and total reward.   
We use Monte Carlo simulation to compute them 
repeating 120 rounds.  

\subsection{Experimental Results}
\begin{figure}[htb]
  \centering
\subfigure[Regret of Algo. \ref{algo:sumup}]{
  \includegraphics[width = .33\textwidth]{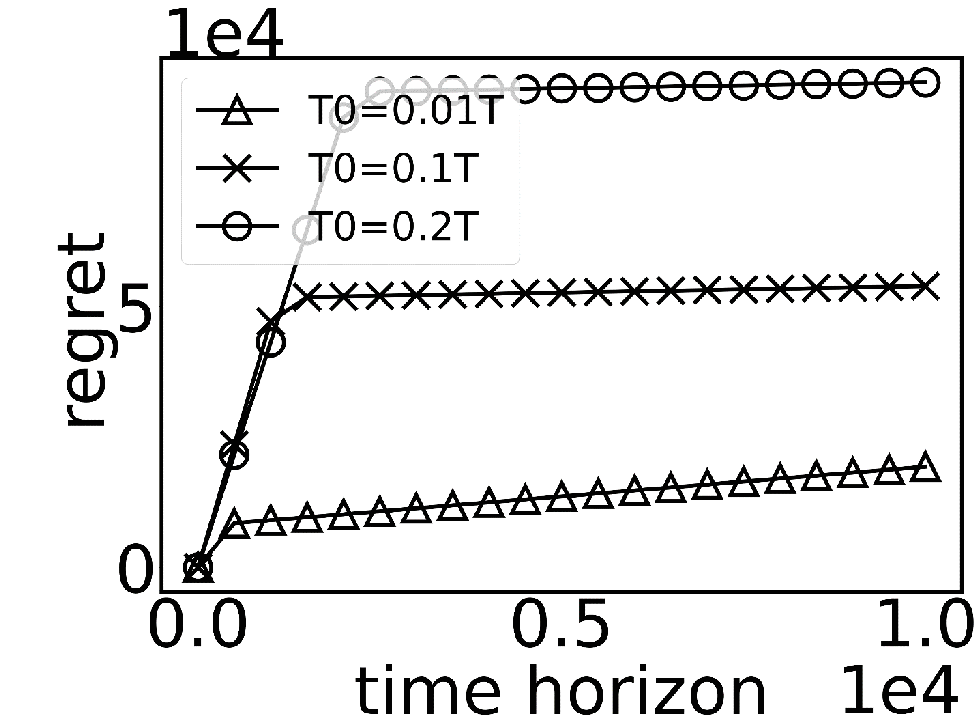}
  \label{fig:RegTime}
}
\subfigure[$T_0=0.01T$]{
  \includegraphics[width = .33\textwidth]{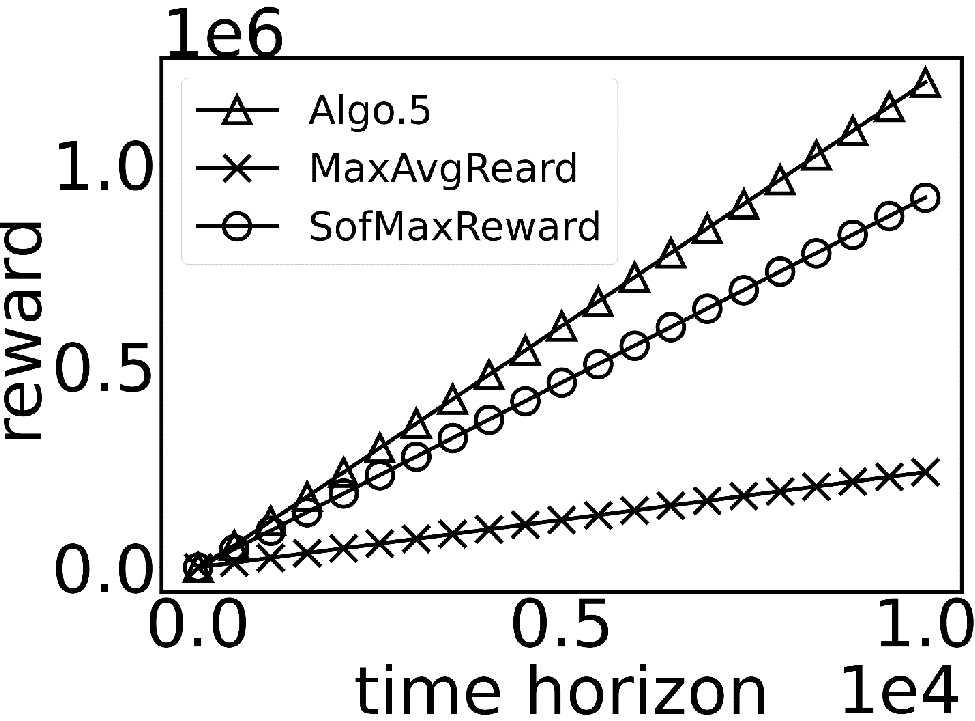}
  \label{fig:focusOnTime001}
}
\\
\subfigure[$T_0=0.1T$]{
  \includegraphics[width = .33\textwidth]{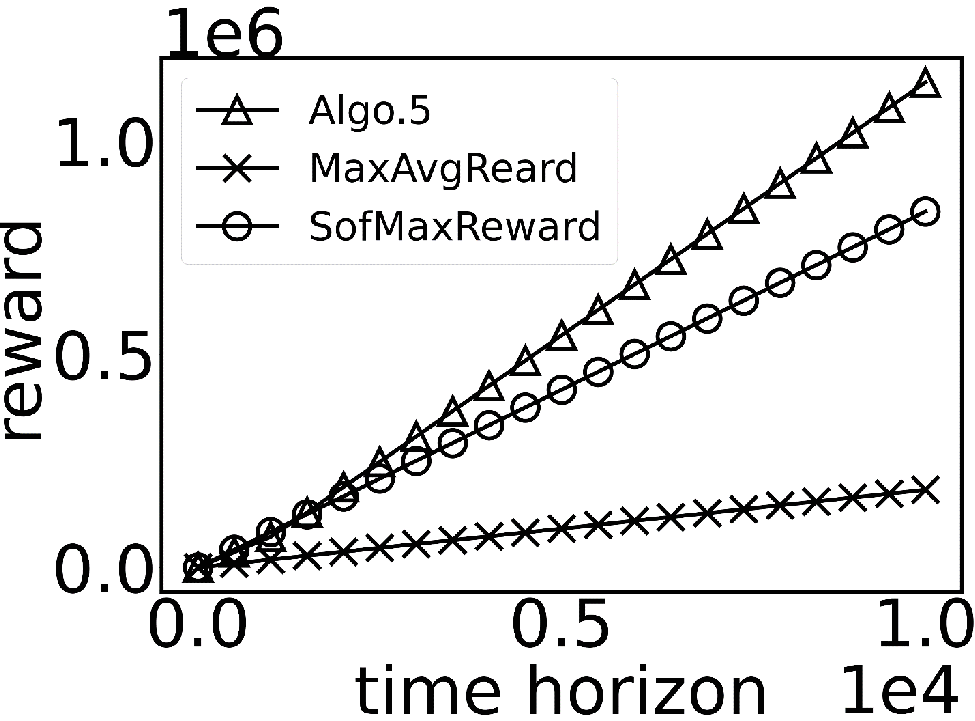}
  \label{fig:focusOnTime01}
}
\subfigure[$T_0=0.2T$]{
  \includegraphics[width = .33\textwidth]{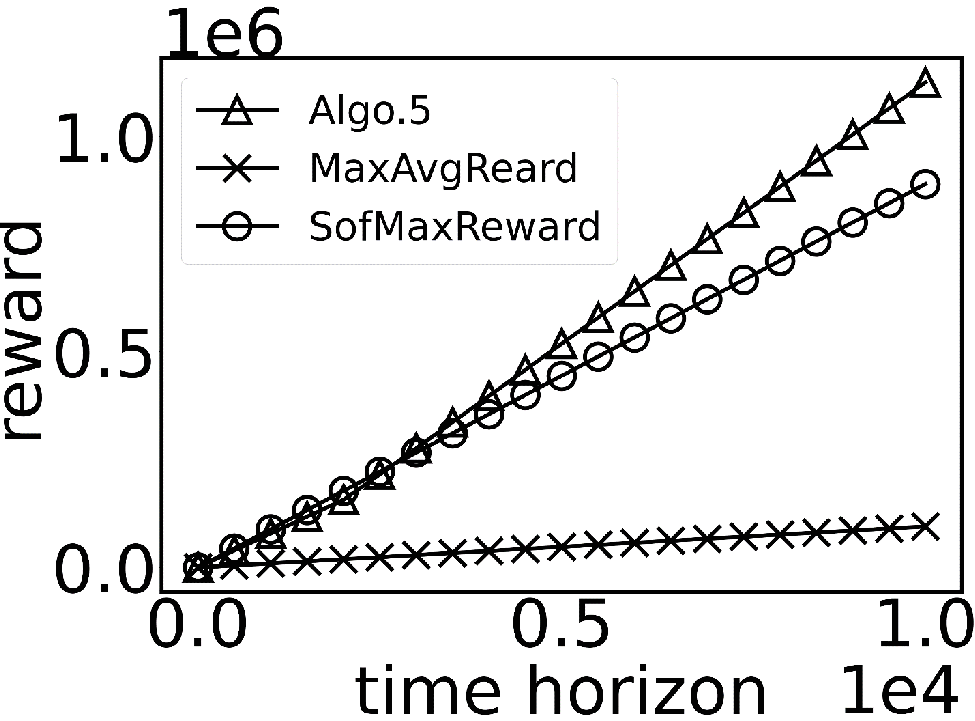}
  \label{fig:focusOnTime02}
}
  \vspace{-0.18in}
\caption{Impact of length of exploration. }
\label{fig:focusOnTime}
  \vspace{-0.18in}
\end{figure}

\noindent
{\bf Impact of exploration}. 
Fig. \ref{fig:RegTime} shows the regret of 
Algorithm \ref{algo:sumup} as we vary the length of 
exploration from $T_0 {=} 0.01 T$ to $T_0 {=} 0.2 T$.  
One can observe that the regret curve first increases 
sharply in the exploration phase, 
and then becomes flat in the committing phase.   
This verifies that Algo. \ref{algo:sumup} 
has a nice convergence property.    
Figure \ref{fig:focusOnTime001} shows that when 
$T_0 {=} 0.1 T$
the reward curve of Algo. \ref{algo:sumup} lies in the top.  
Namely, Algo. \ref{algo:sumup} has a larger reward 
than two comparison baselines.  
This statement also holds  when the 
length of exploration increases as 
shown in Fig. \ref{fig:focusOnTime01} and \ref{fig:focusOnTime02}.  

\begin{figure}[htb]
  \centering
\subfigure[Regret in different number of arms.]{
  \includegraphics[width = .33\textwidth]{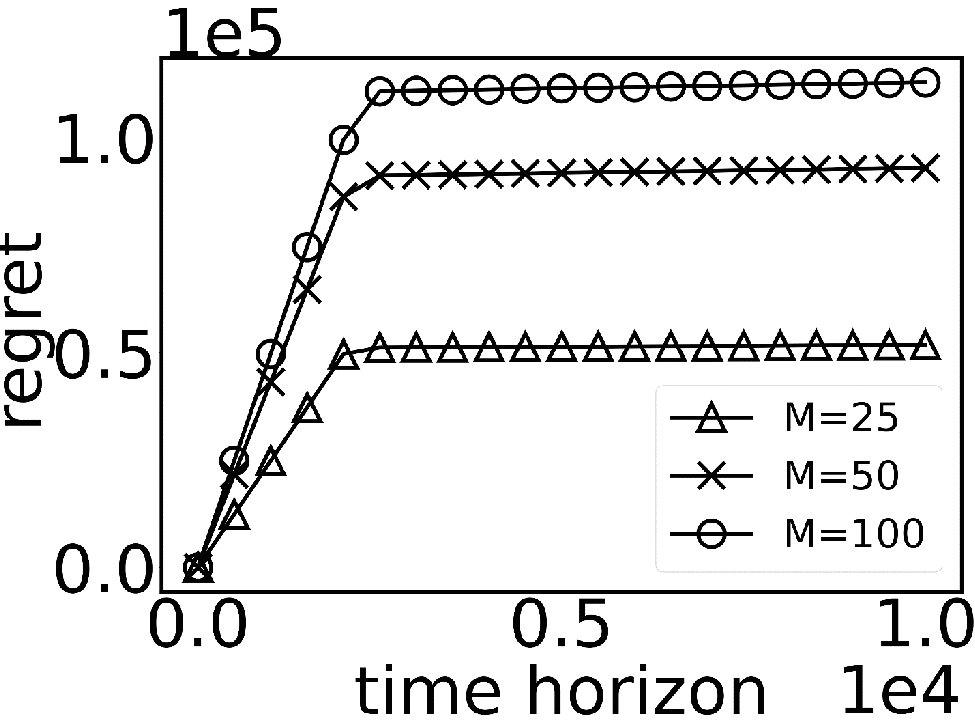}
  \label{fig:RegArm}
}
\subfigure[$M=25$]{
  \includegraphics[width = .33\textwidth]{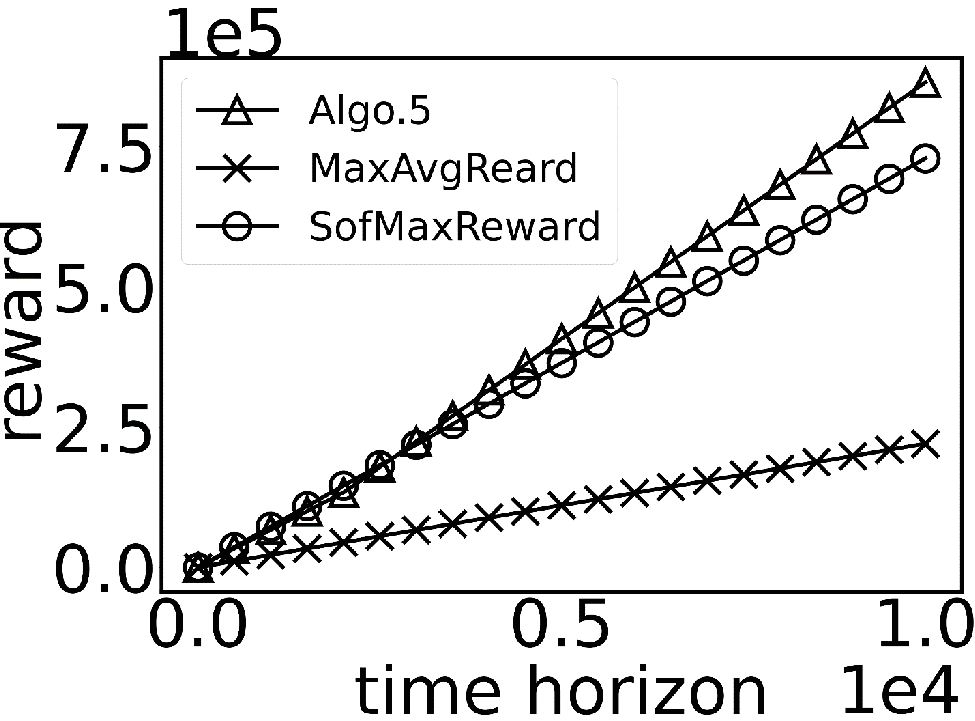}
  \label{fig:focusOnArms25}
}
\subfigure[$M=50$]{
  \includegraphics[width = .33\textwidth]{pic/decay02_Arms50_Players150_variance_01.eps}
  \label{fig:focusOnArms50}
}
\subfigure[$M=100$]{
  \includegraphics[width = .33\textwidth]{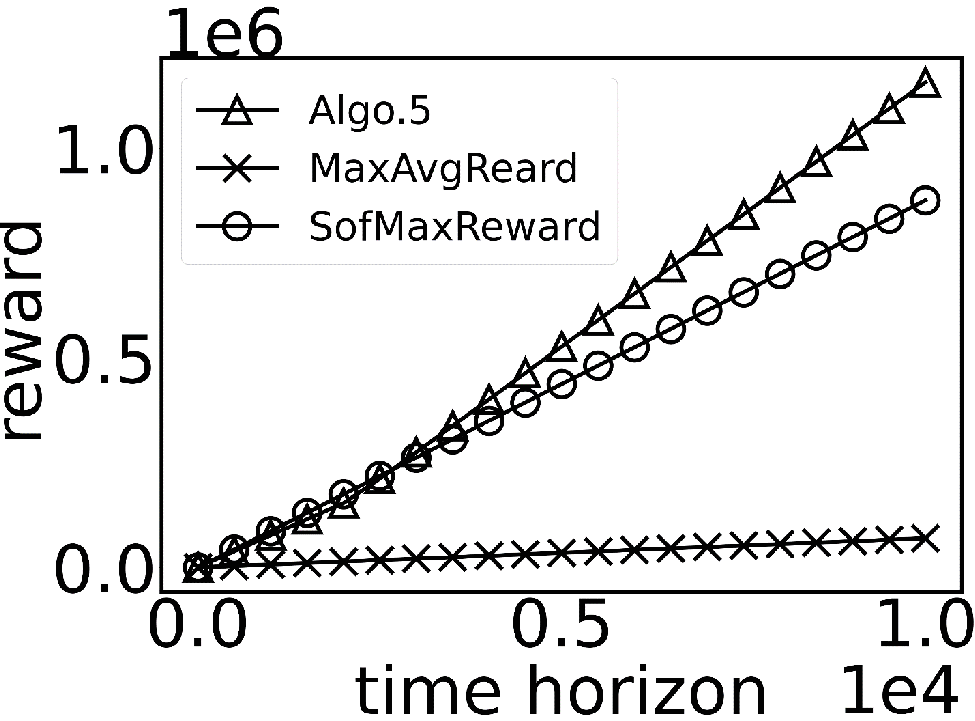}
  \label{fig:focusOnArms100}
}
\caption{Impact on number of arms.}
\label{fig:focusOnArms}
\end{figure}

\begin{figure}[htb]
  \centering
\subfigure[Regret of Algo. \ref{algo:sumup}]{
  \includegraphics[width = .33\textwidth]{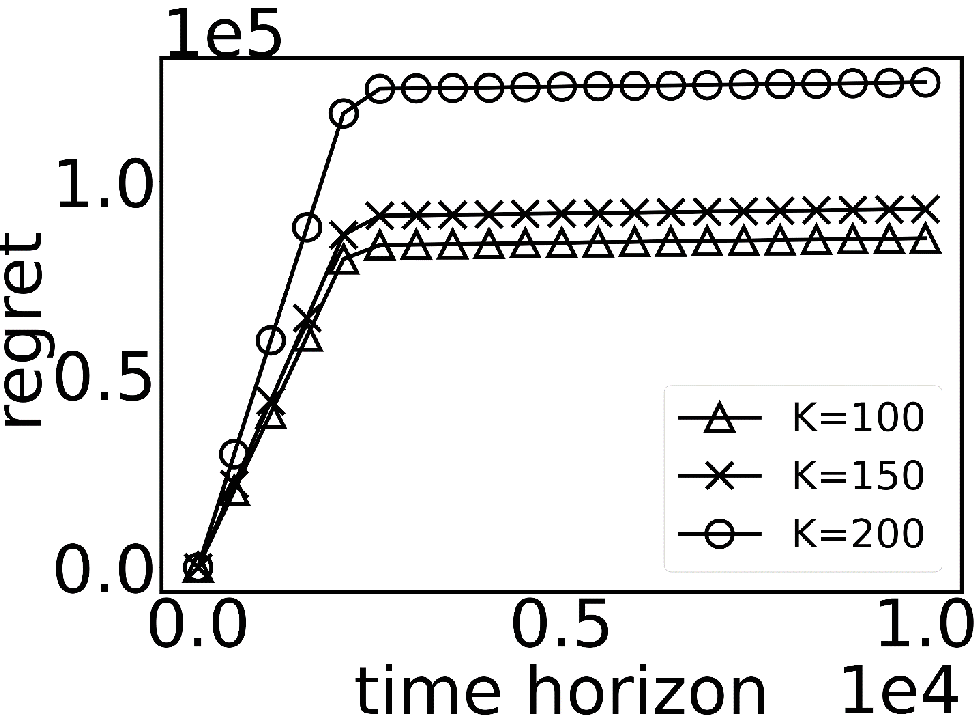}
  \label{fig:RegPlayer}
}
\subfigure[$K=100$]{
  \includegraphics[width = .33\textwidth]{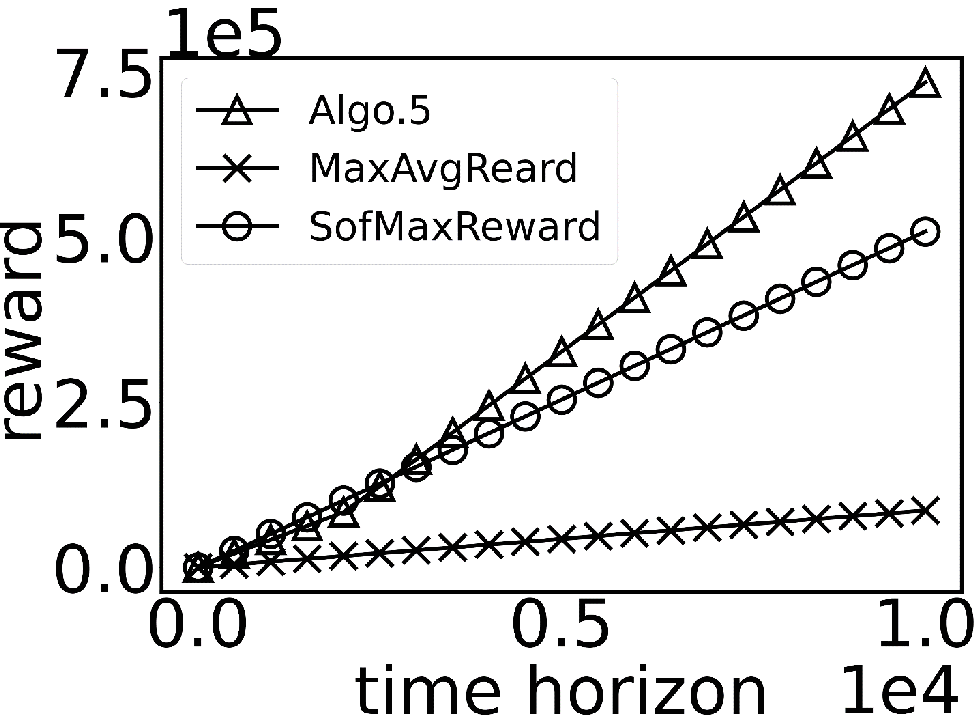}
  \label{fig:focusOnPlayers100}
}
\subfigure[$K=150$]{
  \includegraphics[width = .33\textwidth]{pic/decay02_Arms50_Players150_variance_01.eps}
  \label{fig:focusOnPlayers150}
}
\subfigure[$K=200$]{
  \includegraphics[width = .33\textwidth]{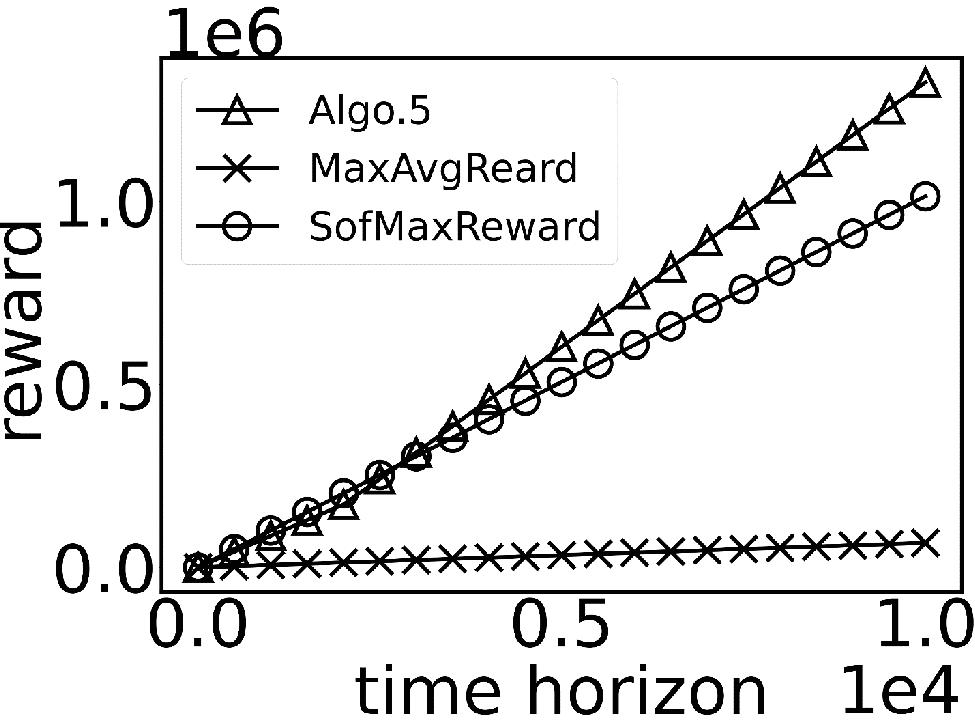}
  \label{fig:focusOnPlayers200}
}
\caption{Impact on number of players. }
\label{fig:focusOnPlayers}
\end{figure}

\noindent
{\bf Impact of number of arms}. 
Figure \ref{fig:RegArm} shows the regret of 
Algorithm \ref{algo:sumup} as we vary the 
number of arms from $M=25$ to $100$. 
From Figure \ref{fig:RegPlayer}, 
one can observe that the regret curves of Algorithm \ref{algo:sumup}  
under different number of arms first increases 
sharply in the exploration phase, 
and then becomes flat in the committing phase.   
This validates that Algorithm \ref{algo:sumup} 
has a nice convergence property under different number of arms.    
Figure \ref{fig:focusOnArms25} shows that 
when $M=25$
the reward curve of Algorithm \ref{algo:sumup} lies in the top.  
Namely, Algorithm \ref{algo:sumup} has a larger reward 
than two comparison baselines.  
This statement also holds as we increase the 
number players as 
shown in Figure \ref{fig:focusOnArms50} and \ref{fig:focusOnArms100}. 
 
\FloatBarrier

\begin{figure}[htb]
  \centering
\subfigure[Regret of Algo. \ref{algo:sumup}]{
  \includegraphics[width = .33\textwidth]{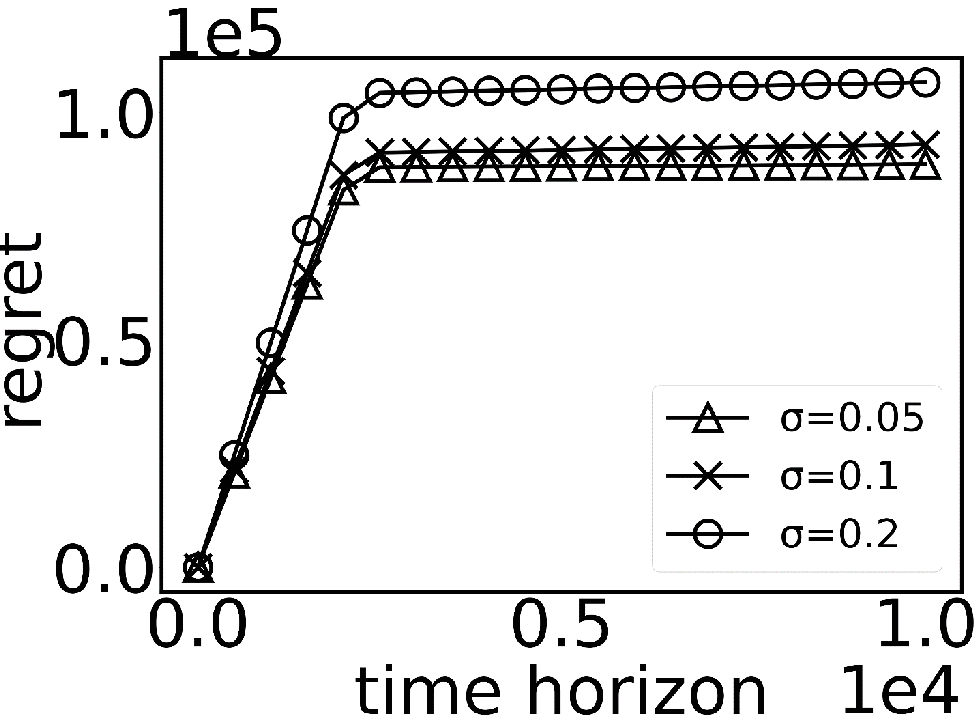}
  \label{fig:RegVariance}
}
\subfigure[$\sigma=0.05$]{
  \includegraphics[width = .33\textwidth]{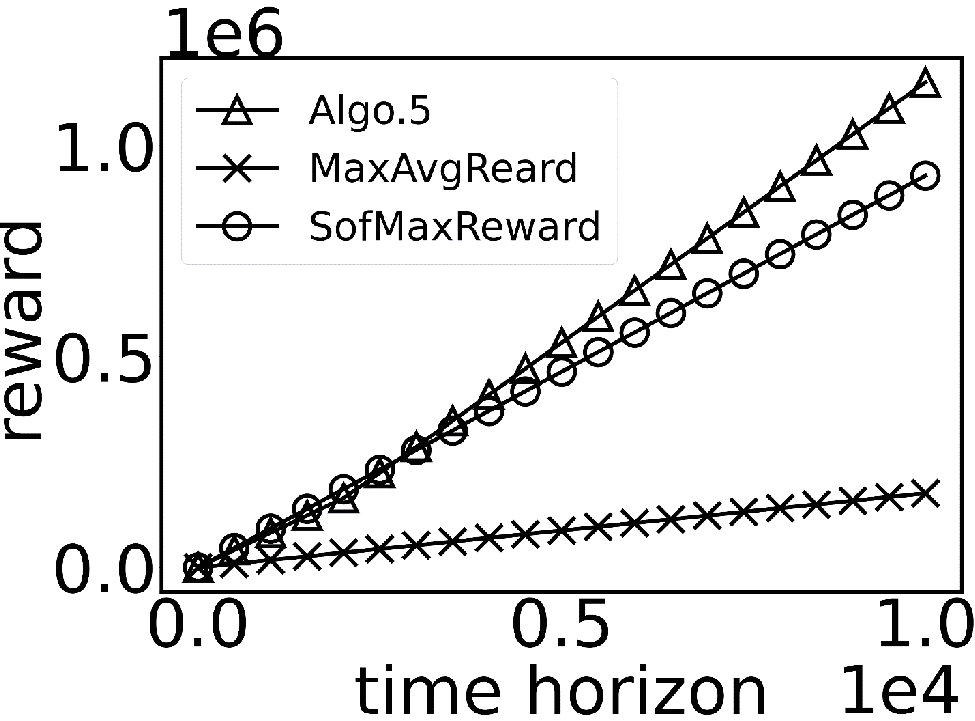}
\label{fig:focusOnVariance05}
}
\subfigure[$\sigma=0.1$]{
  \includegraphics[width = .33\textwidth]{pic/decay02_Arms50_Players150_variance_01.eps}
  \label{fig:focusOnVariance10}
}
\subfigure[$\sigma=0.2$]{
  \includegraphics[width = .33\textwidth]{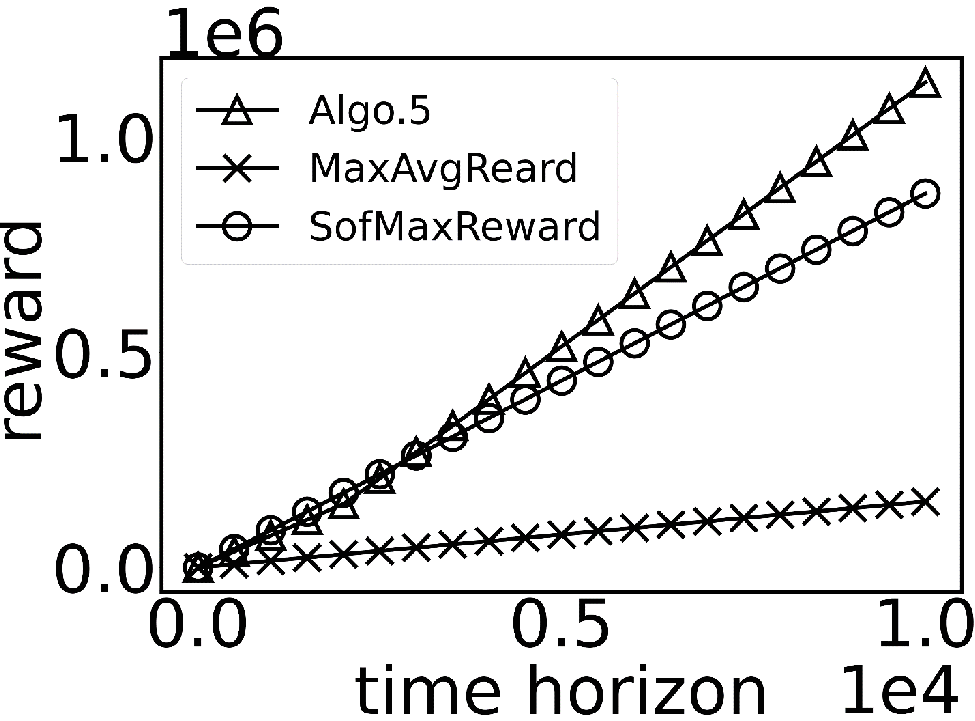}
  \label{fig:focusOnVariance20}
}
\caption{Impact on variance. }
\label{fig:focusOnVariance}
\end{figure}

\noindent
{\bf Impact of number of players}. 
Figure \ref{fig:focusOnPlayers} shows the regret of 
Algorithm \ref{algo:sumup} as we vary the 
number of players from $K=100$ to $200$. 
From Figure \ref{fig:RegPlayer}, 
one can observe that the regret curves of Algorithm \ref{algo:sumup}  
under different number of players first increases 
sharply in the exploration phase, 
and then becomes flat in the committing phase.   
This validates that Algorithm \ref{algo:sumup} 
has a nice convergence property under different number of players.    
Figure \ref{fig:focusOnPlayers100} shows that 
when $K=100$
the reward curve of Algorithm \ref{algo:sumup} lies in the top.  
Namely, Algorithm \ref{algo:sumup} has a larger reward 
than two comparison baselines.  
This statement also holds as we increase the 
number players as 
shown in Figure \ref{fig:focusOnVariance10} and \ref{fig:focusOnVariance20}.

\FloatBarrier

\noindent
{\bf Impact of standard deviation}. 
Figure \ref{fig:RegVariance} shows the regret of 
Algorithm \ref{algo:sumup} as we vary the  
standard deviation from $\sigma=0.05$ to $\sigma=0.2$. 
From Figure \ref{fig:RegVariance}, 
one can observe that the regret curves of Algorithm \ref{algo:sumup} 
under different standard deviations first increases 
sharply in the exploration phase, 
and then becomes flat in the committing phase.   
This validates that Algorithm \ref{algo:sumup} 
has a nice convergence property under different standard deviations.    
Figure \ref{fig:focusOnVariance05} shows that 
when $\sigma=0.05$
the reward curve of Algorithm \ref{algo:sumup} lies in the top.  
Namely, Algorithm \ref{algo:sumup} has a larger reward 
than two comparison baselines.  
This statement also holds as we increase the 
standard deviation as 
shown in Figure \ref{fig:focusOnVariance10} and \ref{fig:focusOnVariance20}.

\FloatBarrier

\section{Conclusion}
This paper formulates a new variant of multi-player 
MAB model for distributed selection problems.  
We designed a computational efficient greedy algorithm, 
to located one of the optimal 
arm pulling profiles.  
We also designed an iterative distributed algorithm for players 
to commit to an optimal arm pulling profile 
with a constant number of rounds in expectation.   
We designed an exploration strategy with a length 
such that each player can have an accurate estimate on the optimal arm pulling profile   
with high probability.   
Such estimates can be different across different players.  
We designed an iterative distributed algorithm, 
which guarantees that players arrive at a consensus on the 
optimal arm pulling profile.  
We conduct experiments to validate our algorithms.    

%
%
%
\bibliographystyle{splncs04}
%
\bibliography{reference} 


\appendix

\section{Technical Proofs}

\subsection{Proof of Lemma \ref{lem:offline:maringalgain}} 

Without loss of generality, let us consider an arbitrary arm $m \in \mathcal{M}$.  
First, the $U_m(n_{t,m},\bm{p}_m,\mu_m)$ can be derived as:  
\begin{align*}
U_m(n_{t,m},
\bm{p}_m,
\mu_m) 
&= \mu_m
\sum_{d\in\mathcal{D}}  p_{m,d}  \min\{n_{t,m},d\} \\
&= \mu_m  \sum_{d=1}^{n_{t,m}-1}  dp_{m,d}
+  \mu_m  n_{t,m}  \left( 1-\sum_{d=1}^{n_{t,m}-1}p_{m,d}  \right)
\end{align*} 
Based on this analytical expression of $U_m(n_{t,m},\bm{p}_m,\mu_m)$, 
the $\Delta_m(n)$ can be derived as:  
\begin{align*}
\Delta_m(n)
&=
U_m(n+1,\bm{p}_m,\mu_m)  -    U_m(n,\bm{p}_m,\mu_m)  \\
&=\mu_m\left[
\sum_{d=1}^{n}    dp_{m,d}  +   (n+1) \left(  1-\sum_{d=1}^{n}p_{m,d}   \right)
-  \sum_{d=1}^{n-1}  dp_{m,d}
-   n\left(1-\sum_{d=1}^{n-1}p_{m,d}\right)
\right]\\
&=\mu_m\left[
np_{m,n}
+1-np_{m,n}
-\sum_{d=1}^{n}p_{m,d}
\right]\\
&=\mu_m   \left(  1-\sum_{d=1}^{n}  p_{m,d}  \right) \\
&=\mu_m   \sum_{d=n+1}^{d_{\max}}   p_{m,d}
\end{align*} 
Then, it follows that 
\begin{equation*}
\Delta_m(n+1)  -    \Delta_m(n)
=
\mu_m \left[   \sum_{d=n+2}^{d_{\max}}p_{m,d}
-\sum_{d=n+1}^{d_{\max}}p_{m,d}   \right]
=
- \mu_mp_{m,n+1}
\leq 
0,  
\end{equation*}
where the last step holds 
because the probability $p_{m,n+1} \geq 0, \forall d\in\mathcal{D} $.  
This proof is then complete.  

\subsection{Proof of Theorem \ref{thm:offline:GreedyGuarantee}}

Consider an arbitrary assignment of players to arms 
$\bm{n} = [n_1, \ldots, n_M]$.  
\begin{align*}
U \left(\bm{n},\bm{P}, \bm{\mu}\right) 
& 
= 
\sum_{m=1}^{M} U_m (n_m, \bm{p}_m, \mu_m) 
\\
& 
= \sum_{m=1}^{M} 
\left( 
U_m (0, \bm{p}_m, \mu_m) 
+
\sum^{n_m}_{n=1} 
( U_m (n, \bm{p}_m, \mu_m)  
- 
U_m (n-1, \bm{p}_m, \mu_m)
)  
\right) 
\\
& 
= \sum_{m=1}^{M} 
\left( 
U_m (0, \bm{p}_m, \mu_m) 
+
\sum^{n_m}_{n=1} 
\Delta_m (n-1)
\right) 
\\
& 
= \sum_{m=1}^{M}  
\sum^{n_m}_{n=1} 
\Delta_m (n-1),    
\end{align*}
where the last step is due to that 
$U(0, \bm{p}_m, \mu_m)  = \mu_m \min\{0,X_i\}=0, \forall m\in\mathcal{M}$.  
Thus, each optimal arm pulling profile should corresponds to 
top-$K$ elements in the vector 
$(\Delta_m(n): m\in \mathcal{M}, n \in \{0, \ldots, K-1\})$.  
Lemma \ref{lem:offline:maringalgain} states that for each arm $m$, 
$\Delta_m(n)$ is non-increasing in $n$.  
Hence, Algorithm \ref{algo:greedy} selects 
top-$K$ elements in the vector 
$(\Delta_m(n): m\in \mathcal{M}, n \in \{0, \ldots, K-1\})$. 
The optimality of the output of Algorithm \ref{algo:greedy} 
is then complete.  

Now we analyze the computational complexity of Algorithm \ref{algo:greedy}. 
In total there are $K$ loops. 
In each loop, we must find the biggest marginal gain from $M$ arms.   
This can be achieved by using sorting algorithm such as merge sort. 
Consider a sorting algorithm with computational complexity $O(M\ln(M))$.  
Then. we conclude the computational complexity of 
Algorithm \ref{algo:greedy}.  

\subsection{An Technical Lemma}

\begin{lemma}
The conditional expectation  
$
\mathbb{E}
[
\bm{I}(
n^-_{t+1, m} 
= 0 
)
| 
N^-_t, 
n^-_{t, m}
]
$
can be derived as: 
\begin{equation}
\mathbb{E}
[
\bm{I}(
n^-_{t+1, m} 
= 0 
)
| 
N^-_t, 
n^-_{t, m}
]
=
{
N^-_t 
\choose
n^-_{t, m}
}
p_{t,m}^{ n^-_{t, m} } 
(1-p_{t,m})^{N^-_t- n^-_{t, m}}, 
\label{eq:offline:commitProb}
\end{equation}
where $\bm{I}(\cdot)$ is an indicator function and  
$p_{t,m} 
\triangleq 
n^-_{t, m} 
/ 
N^{-}_t$.    
The 
conditional expectation 
$
\mathbb{E}
[
\bm{I}(
n^-_{t+1, m} 
= 0 
)
| 
N^-_t, 
n^-_{t, m}
]
$
is nondecreasing in $t$.  
Furthermore, we have 
\[
\mathbb{P}
[
n^-_{t+1, m} 
> 
0
]
\leq 
\left(
1 
- 
{
K 
\choose
n^\ast_{m}
}
\left(
\frac{n^\ast_{m}}{K}
\right)^{ n^\ast_{m} } 
\left(
1
-
\frac{n^\ast_{m}}{K}
\right)^{K- n^\ast_{m} }
\right)^{t+1}.  
\]
\label{lem:commitArm}
\end{lemma}

{\bf Proof: }
For notation simplicity we omit the stating time slots $T_{\text{start}}$.  
The following results hold by directly adding $T_{\text{start}}$ 
to $t$.   
Recall that $N_{t}^-$ denotes the total number of 
uncommitted players up to round $t$ 
and $n^-_{t,m}$ denotes the number of players that 
arm $m$ lacks up to round $t$.  
It would make sense that $N_{t}^->1$ and $n^-_{t,m}>1$.  
Note that in round $t+1$, 
each uncommitted player pulls to arm $m$ with probability 
$p_{t,m} \triangleq n^-_{t,m} / N_{t}^-$.  
In round $t+1$, arm $m$ becomes fully committed 
if and only if $n_{t+1,m}^- = 0$.  
Note that $n_{t+1,m}^- = 0$ is equivalent to 
that $n^-_{t,m}$ players out of $N_{t}^-$ players pull arm $m$.  
Hence, we have:  
\begin{align}
\mathbb{E}
[
\bm{I}(
n^-_{t+1, m} 
= 0 
)
| 
N^-_t, 
n^-_{t, m}
]
& 
=
{N_{t}^- 
\choose 
n_{t,m}^-
} 
p_{t,m}^{n^-_{t, m}}
(1-p_{t,m})^{N_{t}^-  - n^-_{t, m}}
\nonumber
\\
& 
={N_{t}^- 
\choose 
n_{t,m}^-
} 
\left(
\frac{n_{t,m}^-}{N_{t}^- }
\right)^{n^-_{t, m}}
\left(
1
-
\frac{n_{t,m}^-}{N_{t}^- }
\right)^{N_{t}^-  - n^-_{t, m}}.  
\label{eq:appendix:committedBase}
\end{align} 

To conclude the monotonicity property, 
we prove 
$
\mathbb{E}
[
\bm{I}(
n^-_{t+1, m} 
= 0 
)
| 
N^-_t, 
n^-_{t, m}
]
\leq 
\mathbb{E}
[
\bm{I}(
n^-_{t+2, m} 
= 0 
)
| 
N^-_{t+1}, 
n^-_{t+1, m}
]
$.  
First, Equation (\ref{eq:appendix:committedBase}) implies that 
\begin{align}
& 
\mathbb{E}
[
\bm{I}(
n^-_{t+2, m} 
= 0 
)
| 
N^-_{t+1}, 
n^-_{t+1, m}
]
\\
& 
={N_{t+1}^- 
\choose 
n_{t+1,m}^-
} 
\left(
\frac{n_{t+1,m}^-}{N_{t+1}^- }
\right)^{n^-_{t+1, m}}
\left(
1
-
\frac{n_{t+1,m}^-}{N_{t+1}^- }
\right)^{N_{t+1}^-  - n^-_{t+1, m}}.  
\label{eq:appendix:committedINduction}
\end{align} 
Note that $n_{t+1,m}^- \leq n_{t,m}^-$.  
Then it follows that 
\[
N_{t+1}^- 
= 
\sum_{m \in \mathcal{M}} 
n_{t+1,m}^-
\leq 
\sum_{m \in \mathcal{M}} 
n_{t,m}^-
=
N_{t}^- .  
\] 
Thus, it suffices to prove that 
$
\mathbb{E}
[
\bm{I}(
n^-_{t+1, m} 
= 0 
)
| 
N^-_t, 
n^-_{t, m}
]
\leq 
\mathbb{E}
[
\bm{I}(
n^-_{t+2, m} 
= 0 
)
| 
N^-_{t+1}, 
n^-_{t+1, m}
]
$
holds under two cases: 
(1) $N_{t+1}^- = N_t^- -1$, 
$n_{t+1,m}^- =n^-_{t,m}$; 
(2) $N_{t+1}^- = N_t^- -1$, 
$n_{t+1,m}^- = n^-_{t,m}-1$.   
One can use these two cases as basis and use induction 
to conclude the general case when the difference  
$n_{t+1,m}^- - n^-_{t,m}$ and 
$N_{t+1}^- - N_t^-$ are bigger than one.  
We proceed to consider the first case $N_{t+1}^- = N_t^- -1$, 
$n_{t+1,m}^- =n^-_{t,m}$.  
Plugging $N_{t+1}^- = N_t^- -1$ and  
$n_{t+1,m}^- =n^-_{t,m}$
into Equation (\ref{eq:appendix:committedINduction}) we have that 
\begin{align*}
& 
\mathbb{E}
[
\bm{I}(
n^-_{t+2, m} 
= 0 
)
| 
N^-_{t+1}, 
n^-_{t+1, m}
]
\\
& 
=
{
N^-_t - 1
\choose
n^-_{t, m}
}
\left(
\frac{n_{t,m}^-}
{N_t^- -1}
\right)^{n^-_{t, m}}
\left(
1-
\frac{n_{t,m}^-}
{N_t^- -1}
\right)^{N^-_t - 1-n^-_{t, m}}.  
\end{align*}
Comparing the above equation with Equation (\ref{eq:appendix:committedBase}), 
and by rearranging terms, we have 
\begin{align}
& 
\mathbb{E}
[
\bm{I}(
n^-_{t+1, m} 
= 0 
)
| 
N^-_t, 
n^-_{t, m}
]
\nonumber
\\
&
=
\left(
\frac{N^-_t   -   n^-_{t, m}}{N^-_t -1 -n^-_{t, m}}
\right)^{N^-_t - 1   - n^-_{t, m}  }
\left(   \frac{N^-_t - 1}  {N^-_t }   \right) ^ {N^-_t - 1}
\mathbb{E}
[
\bm{I}(
n^-_{t+2, m} 
= 0 
)
| 
N^-_{t+1}, 
n^-_{t+1, m}
].  
\end{align} 
Then it suffices to show  
\begin{align*}
& 
\left(
\frac{N^-_t   -   n^-_{t, m}}{N^-_t -1 -n^-_{t, m}}
\right)^{N^-_t - 1   - n^-_{t, m}  }
\left(   \frac{N^-_t - 1}  {N^-_t }   \right) ^ {N^-_t - 1}
\leq 
1
\\
& 
\Leftrightarrow
\left(
\frac{N^-_t   -   n^-_{t, m}}{N^-_t -1 -n^-_{t, m}}
\right)^{N^-_t - 1   - n^-_{t, m}  }
\leq 
\left( \frac{N^-_t }{N^-_t - 1} \right) ^ {N^-_t - 1}
\\
& 
\Leftrightarrow
\left(
1 + 
\frac{
1  
}{ N^-_t -1 -n^-_{t, m} }
\right)^{ N^-_t - 1   - n^-_{t, m} }
\leq 
\left( 1 + \frac{ 1 }{N^-_t - 1} \right) ^ {N^-_t - 1}.  
\end{align*}
A sufficient condition to assert the above inequality is that 
$(1 + \frac{1}{x})^x$ is increasing in $x \geq 1$.  
To achieve this, 
denote $f(x)=\ln{\left(\frac{1+x}{x}\right)^x}=x\left[\ln{(1+x)}-\ln{x}\right]$.  
Taking the first order derivative of $f(x)$, 
we have 
\[
f'(x)=\frac{df(x)}{dx}=\ln{\frac{1+x}{x}}-\frac{1}{1+x}.  
\] 
Further taking the first order derivative of $f'(x)$, we have 
\[
f''(x)=\frac{df'(x)}{dx}=
\frac{1}{\left(1+x\right)^2}
-\frac{1}{(1+x)x}<0,  
\]
where the last inequality holds because we focus on the case 
that $x\geq 1$.  
This implies that $f'(x)$ is decreasing in $x\geq 1$.   
Also note that $\lim_{x \rightarrow +\infty} f'(x) = 0$.  
Hence, we have that $f'(x) > 0$.  
This implies that $f(x)$ is increasing in $x\geq1$.  
Namely, $(1 + \frac{1}{x})^x$ is increasing in $x \geq 1$.  
We therefore conclude the first case.  
Now we consider the second case  
$N_{t+1}^-
=N_t^- -1$, 
$n_{t+1,m}^-
 =n^-_{t,m}-1$.   
Plugging $N_{t+1}^-
=N_t^- -1$
and 
$n_{t+1,m}^-
 =n^-_{t,m}-1$
into Equation (\ref{eq:appendix:committedINduction}) we have that  
\begin{equation*}
\mathbb{E}
[
\bm{I}(
n^-_{t+2, m} 
= 0 
)
| 
N^-_{t+1}, 
n^-_{t+1, m}
]
=
{
N^-_t - 1
\choose
n^-_{t, m}-1
}
\left(
\frac{n^-_{t,m}-1}{N_t^- -1}
\right)^{n^-_{t, m}-1}
\left(
1-
\frac{n^-_{t,m}-1}{N_t^- -1}
\right)^{N^-_t -n^-_{t, m}}
\end{equation*}
Comparing the above equation with Equation (\ref{eq:appendix:committedBase}), 
and by rearranging terms, we have 
\begin{equation*}
\mathbb{E}
[
\bm{I}(
n^-_{t+1, m} 
= 0 
)
| 
N^-_t, 
n^-_{t, m}
]
=
\left(
\frac{n^-_{t, m}}{n^-_{t, m}-1}
\right)^{n^-_{t, m}  - 1}
\left(   \frac{N^-_t - 1}  {N^-_t }   \right) ^ {N^-_t - 1}
\mathbb{E}
[
\bm{I}(
n^-_{t+2, m} 
= 0 
)
| 
N^-_{t+1}, 
n^-_{t+1, m}
]
\end{equation*} 
Then, it suffices to show 
\begin{align*}
& 
\left(
\frac{n^-_{t, m}}{n^-_{t, m}-1}
\right)^{n^-_{t, m}  - 1}
\left(   \frac{N^-_t - 1}  {N^-_t }   \right) ^ {N^-_t - 1}
\leq 
1
\\
& 
\Leftrightarrow
\left(
\frac{n^-_{t, m}}{n^-_{t, m}-1}
\right)^{n^-_{t, m}  - 1}
\leq
\left(   \frac {N^-_t } {N^-_t - 1}   \right) ^ {N^-_t - 1} 
\\
& 
\Leftrightarrow
\left(
1 + 
\frac{1}{n^-_{t, m}-1}
\right)^{n^-_{t, m}  - 1}
\leq
\left( 
1 + \frac { 1 } {N^-_t - 1} \right) ^ {N^-_t - 1} 
\end{align*}
Note that $n^-_{t, m} \leq N^-_t$.  
A sufficient condition to assert the above inequality is that 
$(1 + \frac{1}{x})^x$ is increasing in $x \geq 1$.   
This sufficient condition has been proved in the proof of the first case.  
We therefore conclude the second case.  
In summary, we conclude that the probability 
$\mathbb{E}
[
\bm{I}(
n^-_{t+1, m} 
= 0 
)
| 
N^-_t, 
n^-_{t, m}
]$ 
is non-decreasing in $t$.  
Hence, an upper bound of 
$\mathbb{E}
[
\bm{I}(
n^-_{t+1, m} 
= 0 
)
| 
N^-_t, 
n^-_{t, m}
]$ can be derived as 
\[
\mathbb{E}
[
\bm{I}(
n^-_{t+1, m} 
= 0 
)
| 
N^-_t, 
n^-_{t, m}
]
\geq
\mathbb{E}
[
\bm{I}(
n^-_{1, m} 
= 0 
)
| 
N^-_0, 
n^-_{0, m}
]
=
{
K 
\choose
n^\ast_{m}
}
\left(
\frac{n^\ast_{m}}{K}
\right)^{ n^\ast_{m} } 
\left(
1
-
\frac{n^\ast_{m}}{K}
\right)^{K- n^\ast_{m} }.  
\]
Note that by the two probability of conditional expectation, we have 
\begin{align*}
\mathbb{E}
[
\bm{I}(
n^-_{t+1, m} 
= 0 
)
|   
n^-_{t, m}
]
& 
= 
\mathbb{E}
[
\mathbb{E}
[
\bm{I}(
n^-_{t+1, m} 
= 0 
)
| 
N^-_t, 
n^-_{t, m}
] 
| 
n^-_{t, m} 
]
\\
& 
\geq 
{
K 
\choose
n^\ast_{m}
}
\left(
\frac{n^\ast_{m}}{K}
\right)^{ n^\ast_{m} } 
\left(
1
-
\frac{n^\ast_{m}}{K}
\right)^{K- n^\ast_{m} }. 
\end{align*}
The above inequality implies that for any $n$ satisfying $\mathbb{P}[n^-_{t, m} 
= n] >0$, it holds that 
\[
\mathbb{E}
[
\bm{I}(
n^-_{t+1, m} 
= 0 
)
|   
n^-_{t, m} 
= n
]
\geq 
{
K 
\choose
n^\ast_{m}
}
\left(
\frac{n^\ast_{m}}{K}
\right)^{ n^\ast_{m} } 
\left(
1
-
\frac{n^\ast_{m}}{K}
\right)^{K- n^\ast_{m} }. 
\]
Note that 
$
\mathbb{P}
[ 
n^-_{t+1, m} 
= 0 
|   
n^-_{t, m} 
= n
]
= 
\mathbb{E}
[
\bm{I}(
n^-_{t+1, m} 
= 0 
)
|   
n^-_{t, m} 
= n
]
$.
Then it follows that  
\[
\mathbb{P}
[ 
n^-_{t+1, m} 
= 0 
|   
n^-_{t, m} 
= n
]
\geq 
{
K 
\choose
n^\ast_{m}
}
\left(
\frac{n^\ast_{m}}{K}
\right)^{ n^\ast_{m} } 
\left(
1
-
\frac{n^\ast_{m}}{K}
\right)^{K- n^\ast_{m} }. 
\]

Note that an arm is not fully committed up to round $t+1$, 
if and only if 
$n^-_{t+1, m} 
> 
0$.  
Note that $n^-_{t, m}$ is non-increasing in $t$.  
Hence, we have that 
$n^-_{t+1, m} 
> 
0$ 
is equivalent to that 
arm $m$ is not fully committed in round $1$ to $t+1$, 
i.e, $n^-_{1, m} 
> 
0, 
\ldots, n^-_{t+1, m} 
> 
0$.  
By the chain rule we have that  
\[
\mathbb{P} [ n^-_{t+1, m} > 0  ] 
=
\bigcap_{s=1}^{t+1} 
\mathbb{P} [ n^-_{s, m} > 0  | n^-_{s-1, m} > 0].  
\]
We conclude this theorem by showing the following: 
\begin{align*}
& 
\mathbb{P}
[n_{s,m}^- > 0 | n_{s-1,m}^- > 0 ] 
\\
& 
= 
\frac{ 
\mathbb{P}
[n_{s,m}^- > 0, n_{s-1,m}^- > 0 ] 
}{
\mathbb{P}
[n_{s-1,m}^- > 0 ] 
}
\\
& 
=
\frac{ 
\mathbb{P}
[n_{s,m}^- > 0, n_{s-1,m}^- > 0 ] 
+ 
\mathbb{P}
[n_{s,m}^- > 0, n_{s-1,m}^- \leq 0 ] 
}{
\mathbb{P}
[n_{s-1,m}^- > 0 ] 
}
\\
& 
=
\frac{ 
\mathbb{P}
[n_{s,m}^- > 0] 
}{
\mathbb{P}
[n_{s-1,m}^- > 0 ] 
}
\\
& 
=
\frac{ 
\sum_{ n} 
\mathbb{P}
[n_{s-1,m}^- = n] 
\mathbb{P}
[n_{s,m}^- > 0 | n_{s-1,m}^- = n] 
}{
\mathbb{P}
[n_{s-1,m}^- > 0 ] 
}
\\
& 
= 
\frac{ 
\mathbb{P}
[n_{s-1,m}^- =0] 
\mathbb{P}
[n_{s,m}^- > 0 | n_{s-1,m}^- = 0]  
+ 
\sum_{ n\ > 0} 
\mathbb{P}
[n_{s-1,m}^- = n] 
\mathbb{P}
[n_{s,m}^- > 0 | n_{s-1,m}^- = n] 
}{
\mathbb{P}
[n_{s-1,m}^- > 0 ] 
}
\\
& 
= 
\frac{ 
\sum_{n >  0} 
\mathbb{P}
[n_{s-1,m}^- = n] 
\mathbb{P}
[n_{s,m}^- > 0 | n_{s-1,m}^- = n] 
}{
\mathbb{P}
[n_{s-1,m}^- > 0 ] 
}
\\
\\
& 
= 
\frac{ 
\sum_{n >  0} 
\mathbb{P}
[n_{s-1,m}^- = n] 
(1 - \mathbb{P}
[n_{s,m}^- = 0 | n_{s-1,m}^- = n] 
)
}{
\mathbb{P}
[n_{s-1,m}^- > 0 ] 
}
\\
& 
\leq 
\frac{ 
\sum_{n >  0} 
\mathbb{P}
[n_{s-1,m}^- = n] 
}{
\mathbb{P}
[n_{s-1,m}^- > 0 ] 
}
(
1- 
{
K 
\choose
n^\ast_{m}
}
\left(
\frac{n^\ast_{m}}{K}
\right)^{ n^\ast_{m} } 
\left(
1
-
\frac{n^\ast_{m}}{K}
\right)^{K- n^\ast_{m} }
)
\\
& 
= 
1- 
{
K 
\choose
n^\ast_{m}
}
\left(
\frac{n^\ast_{m}}{K}
\right)^{ n^\ast_{m} } 
\left(
1
-
\frac{n^\ast_{m}}{K}
\right)^{K- n^\ast_{m} }.
\end{align*}


\subsection{Proof of Theorem \ref{thm:offline:ConvergenceAlgoComit}}

For notation simplicity we setting the starting time slots 
$T_{\text{start}} =0$.    
Let $T_m$ denote the number of rounds that 
Algorithm \ref{algo:commitOpt} takes make arm $m$ full committed.  
Note that $T_m \geq t$ 
if and only if $n^-_{t-1, m} > 0$.  
By the proof of Lemma \ref{lem:commitArm}, 
we have that 
\[
\mathbb{P}[T_m \geq s] 
=
\mathbb{P}[n^-_{s-1, m} > 0]
\leq 
\left(
1 
- 
{
K 
\choose
n^\ast_{m}
}
\left(
\frac{n^\ast_{m}}{K}
\right)^{ n^\ast_{m} } 
\left(
1
-
\frac{n^\ast_{m}}{K}
\right)^{K- n^\ast_{m} }
\right)^{s-1}.  
\]
Then, the expectation of $\mathbb{E} 
[T_m] $ can be derived as:  
\begin{align*}
\mathbb{E} 
[T_m] 
& 
= 
\sum_{s =1}^\infty 
\mathbb{P}[T_m \geq s]
= 
\sum_{s =1}^\infty 
\mathbb{P}[n^-_{s-1, m} > 0] 
\\
& 
\leq 
\sum_{s =1}^\infty 
\left(
1 
- 
{
K 
\choose
n^\ast_{m}
}
\left(
\frac{n^\ast_{m}}{K}
\right)^{ n^\ast_{m} } 
\left(
1
-
\frac{n^\ast_{m}}{K}
\right)^{K- n^\ast_{m} }
\right)^{s-1} 
\\
& 
= 
\frac{1} 
{ 
{
K 
\choose
n^\ast_{m}
}
\left(
\frac{n^\ast_{m}}{K}
\right)^{ n^\ast_{m} } 
\left(
1
-
\frac{n^\ast_{m}}{K}
\right)^{K- n^\ast_{m} }
}.  
\end{align*}
Let $T_c$ denote the number of rounds that 
Algorithm \ref{algo:commitOpt} takes to terminate.   
Note that  Algorithm \ref{algo:commitOpt} terminates if 
and only if all arms are fully committed, 
i.e., $T_c = \max_{m \in \mathcal{M} } T_m$.   
Then it follows that 
\[
\mathbb{E} [ T_c ] 
=  
\mathbb{E} 
\left[ 
\max_{m \in \mathcal{M} } T_m
\right] 
\leq 
\mathbb{E} 
\left[ 
\sum_{m \in \mathcal{M} } T_m
\right] 
=
\sum_{m \in \mathcal{M} } 
\mathbb{E} 
\left[ 
T_m
\right] 
=
\sum_{m \in \mathcal{M} }  
\frac{1} 
{ 
{
K 
\choose
n^\ast_{m}
}
\left(
\frac{n^\ast_{m}}{K}
\right)^{ n^\ast_{m} } 
\left(
1
-
\frac{n^\ast_{m}}{K}
\right)^{K- n^\ast_{m} }
}.  
\]
This proof is then complete.  

\subsection{Proof of Lemma \ref{lem:Online:exploration}} 

Applying the Dvoretzky-Kiefer-Wolfowitz Inequality (i.e., Theorem 7.5 of \cite{Wasserman2013}), 
for each arm $m \in \mathcal{M}$, 
we have that 
\begin{align*}
\mathbb{P} 
\left[ 
\forall d \in \mathcal{D}, 
\widehat{P}^{(k)}_{m,d}
- 
P_{m,d}
\leq 
\sqrt{
\frac{ \ln \delta_1^{-1} }{ 2 T_0 }
}
\right]
\geq 
1 - \delta_1.  
\end{align*}
Then by union  bound, 
for all arms, we have 
\begin{align*}
\mathbb{P} 
\left[ 
\forall k \in \mathcal{K}, 
m \in \mathcal{M}, 
d \in \mathcal{D}, 
\widehat{P}^{(k)}_{m,d}
- 
P_{m,d}
\leq 
\sqrt{
\frac{ \ln \delta_1^{-1} }{ 2 T_0 }
}
\right]
\geq 
1 - MK\delta_1.  
\end{align*} 
Denote 
$O_{m,k}(t)=\mathbbm{1}\{$ player $k$ receive a reward 
from arm $m$ in round $t\}$.  
Then we have that 
\[
\mathbb{E} 
[
O_{m,k}(t)
]
= 
\frac{1}{M} 
\sum_{d \in \mathcal{D}} 
p_{m,d} 
\sum^{K-1}_{n =0}  
\frac{ {K-1 \choose n} }{2^{K-1}}
\min \left\{ 1, \frac{d}{n} \right\}.  
\]
Using Hoeffding's inequality, for each arm $m \in \mathcal{M}$, 
we have
\begin{align*}
\mathbb{P} 
\left[
\hat{\mu}^{(k)}_m - \mu_m
\leq
\sqrt{  \frac{\ln{\delta_2^{-1}}}{ 2 \sum_{t=0}^{T_0}O_{m,k}(t) }   }
\right]
\geq 
1 - \delta_2.  
\end{align*}
Using Hoeffding's inequality, we can have 
\begin{equation*}
\mathbb{P}
\left[
\sum_{t=0}^{T_0}O_{m,k}(t)
\leq 
\frac{1}{2}T_0\mathbb{E}[O_{m,k}(t)]
\right]
\leq
\exp 
\left(
-\frac{T_0\mathbb{E}[O_{m,k}(t)]}
{8}
\right).  
\end{equation*}
Then, using union bound, we can prove the following:  
\begin{equation*}
\mathbb{P}
\left[
\sum_{t=0}^{T_0}O_{m,k}(t)
\geq 
\frac{1}{2}T_0\mathbb{E}[O_{m,k}(t)], 
\hat{\mu}^{(k)}_m - \mu_m
\leq
\sqrt{  \frac{\ln{\delta_2^{-1}}}{ 2 \sum_{t=0}^{T_0}O_{m,k}(t) }   }
\right]
\geq
1 - 
\exp(
-\frac{T_0\mathbb{E}[O_{m,k}(t)]}
{8}
)
- \delta_2
\end{equation*}
Note that 
\begin{align*}
& 
\sum_{t=0}^{T_0}O_{m,k}(t)
\geq 
\frac{1}{2}T_0\mathbb{E}[O_{m,k}(t)], 
\hat{\mu}^{(k)}_m - \mu_m
\leq
\sqrt{  \frac{\ln{\delta_2^{-1}}}{ 2 \sum_{t=0}^{T_0}O_{m,k}(t) }   } 
\\
& 
\Rightarrow 
\hat{\mu}^{(k)}_m - \mu_m
\leq
\sqrt{  \frac{\ln{\delta_2^{-1}}}{ T_0\mathbb{E}[O_{m,k}(t)] }   }.  
\end{align*}
Then it follows that  
\begin{align*}
\mathbb{P}
\left[
\hat{\mu}^{(k)}_m - \mu_m
\leq
\sqrt{  \frac{\ln{\delta_2^{-1}}}{ T_0\mathbb{E}[O_{m,k}(t)] }   }
\right]
& 
\geq 
\mathbb{P}
\left[
\sum_{t=0}^{T_0}O_{m,k}(t)
\geq 
\frac{1}{2}T_0\mathbb{E}[O_{m,k}(t)], 
\hat{\mu}^{(k)}_m - \mu_m
\leq
\sqrt{  \frac{\ln{\delta_2^{-1}}}{ 2 \sum_{t=0}^{T_0}O_{m,k}(t) }   }
\right] 
\\
&
\geq
1 - 
\exp(
-\frac{T_0\mathbb{E}[O_{m,k}(t)]}
{8}
)
- \delta_2
\end{align*}
Then by union bound, for all arms and all players, we have 
\begin{align*}
\mathbb{P} 
\left[
\forall m \in \mathcal{M}, 
k \in \mathcal{K}, 
\hat{\mu}^{(k)}_m - \mu_m
\leq
\sqrt{  \frac{\ln{\delta_2^{-1}}}{ T_0 \mathbb{E}[O_{m,k}(t)] }   }
\right]
\geq 
1 - K\sum_{m\in \mathcal{M}}\exp(
-\frac{T_0\mathbb{E}[O_{m,k}(t)]}
{8}
)-MK\delta_2.  
\end{align*}
Finally, by union bound and substitute $c_m$ to $\mathbb{E}[O_{m,k}(t)]$ we conclude this lemma.  

\subsection{Proof of Theorem \ref{thm:estOptProfile}}  

Let 
$\epsilon_{m,d} =  
\widehat{P}^{(k)}_{m,d}
- 
P_{m,d}  $ 
and 
$\epsilon^{(k)}_{m} =\widehat{\mu}^{(k)}_m - \mu_m  $.  
A sufficient condition to make 
$
\widehat{\bm{n}}^\ast (k) 
{\in} 
\arg\max_{ \bm{n} \in \mathcal{A} } 
U \left(\bm{n}, \bm{P}, \bm{\mu}\right) 
$ 
hold is that 
$
|
\widehat{\mu}_m^{(k)} \widehat{P}_{m,d}^{(k)}
-
\mu_m P_{m,d}
| 
\leq \gamma / 2
$ 
holds for all $m\in\mathcal{M}$ and $d\in\mathcal{D}$.  
Note that  
\begin{align*}
|
\widehat{\mu}_m^{(k)} \widehat{P}_{m,d}^{(k)}
-
\mu_m P_{m,d}
|
&
=
\left|
(\mu_m+ \epsilon^{(k)}_{m}) (P_{m,d}
+
\epsilon_{m,d})-\mu_m P_{m,d}
\right|\\
&
=
\left|
\epsilon^{(k)}_{m} \hat{P}_{m,d}
+\epsilon_{m,d} \mu_m
+ \epsilon^{(k)}_{m} \epsilon_{m,d}
\right|\\
&
\leq
|\epsilon^{(k)}_{m}|
+|\epsilon_{m,d}|
+|\epsilon^{(k)}_{m}||\epsilon_{m,d}|.  
\end{align*} 
Thus, one refined condition to make 
$
\widehat{\bm{n}}^\ast (k) 
{\in} 
\arg\max_{ \bm{n} \in \mathcal{A} } 
U \left(\bm{n}, \bm{P}, \bm{\mu}\right) 
$ 
hold is that the following inequality holds
\begin{equation}
|\epsilon^{(k)}_{m}|
+|\epsilon_{m,d}|
+|\epsilon^{(k)}_{m}||\epsilon_{m,d}| 
\leq 
\gamma / 2.  
\label{eq:appendix:OptAccuracy}
\end{equation}
With a straightforward extension of the proof of 
Lemma \ref{lem:Online:exploration} we have: 
\begin{align*}
& 
\mathbb{P} 
\left[
\forall k \in \mathcal{K}, 
d \in \mathcal{D}, 
m \in \mathcal{M},  
\left|
\widehat{P}^{(k)}_{m,d}
- 
P_{m,d} 
\right|
\leq 
\sqrt{
\frac{ \ln \delta_1^{-1} }{ 2 T_0 }
}, 
\left| \widehat{\mu}^{(k)}_m - \mu_m \right|
\leq
\sqrt{  \frac{\ln{\delta_2^{-1}}}{ c_m T_0}   }
\right]
\\
& 
\geq 
1 - 2MK\delta_1 - 2MK\delta_2 
- K\sum_{m\in \mathcal{M}}\exp\left(
- T_0 c_m / 8 
\right).   
\end{align*} 
To make Equation (\ref{eq:appendix:OptAccuracy}) holds 
with probability at least 
$
1 - 2MK\delta_1 - 2MK\delta_2 
- KM\exp\left(
- T_0 c_m / 8 
\right)
$
we only need 
\begin{align*} 
|\epsilon^{(k)}_{m}|
+|\epsilon_{m,d}|
+|\epsilon^{(k)}_{m}||\epsilon_{m,d}|
&
\leq
\frac{1}{\sqrt{T_0}}
(\sqrt{
\frac{1}{2}
\ln \delta_1^{-1}
}
+
\sqrt{
\frac{1}{c_m}
\ln \delta_2^{-1}
}
)
+
\frac{1}{T_0}
\frac{1}{\sqrt{2c_m}}
\sqrt{
\ln \delta_1^{-1}
\ln \delta_2^{-1}
}
\\
&
\leq \frac{\gamma}{2}
\end{align*} 
hold for all $m\in\mathcal{M}$ and $d\in\mathcal{D}$
Solve the above inequality with respect to $T_0$, we have 
\begin{equation*}
T_0 = \max_{m\in \mathcal{M}}
\frac{
(2/c_m)
\ln{\delta_1^{-1}}
\ln{\delta_2^{-1}}
}
{
\left(
\sqrt{
0.5\ln{\delta_1^{-1}}
+{c_m}^{-1}
\ln{\delta_2^{-1}}
+(2+2\gamma)
(2c_m)^{-1/2}
\sqrt{\ln{\delta_1^{-1}}\ln{\delta_2^{-1}}}
}
- \sqrt{0.5
\ln{\delta_1^{-1}}}
- \sqrt{c_m^{-1}
\ln{\delta_2^{-1}}}
\right)^2
}.
\end{equation*}
This proof is then complete.  

\subsection{Proof of Lemma \ref{lem:disagree}}

Recall from the proof of Theorem \ref{thm:offline:GreedyGuarantee} that  
$
U \left(\bm{n},\bm{P}, \bm{\mu}\right) 
= \sum_{m=1}^{M}  
\sum^{n_m}_{n=1} 
\Delta_m (n-1)    
$ 
and an optimal arm pulling profile corresponds 
top-$K$ elements in the vector 
$(\Delta_m(n): m\in \mathcal{M}, n \in \{1, \ldots, K-1\})$.  
Hence, two optimal arm pulling profiles only differs in borderline element 
and optimal arm pulling profiles must agree on the elements 
that are larger than the borderline element.  
Note that $p_{m,d} >0$.  
Then, from the proof of Lemma \ref{lem:offline:maringalgain}, 
we obtain that for each arm $m$, 
$\Delta_m (n)$ is decreasing in $n \in \mathcal{K}$.  
Thus, for the $\Delta_m (n)$ each arm $m$, it can have 
at most one borderline element.  
Hence, we conclude that $|n^\ast_m - \widetilde{n}^\ast_m| \leq 1$.  
If $n^\ast_m \neq \widetilde{n}^\ast_m$, 
then  
$\Delta_m ( \max\{n^\ast_m, \widetilde{n}^\ast_m\} )$ 
is a borderline element.  
This proof is then complete.  

\subsection{Proof of Theorem \ref{thm:consensus}}

From Lemma \ref{lem:disagree}, we know that 
different optimal arm pulling profiles only differs in borderline elements.  
And for any two optimal arm pulling profiles, 
we have that $\bm{n}^\ast$ and $\widetilde{\bm{n}}^\ast$, 
$
|n^\ast_m - \widetilde{n}^\ast_m| \leq 1, 
\forall m \in \mathcal{M} 
$
and if $n^\ast_m \neq \widetilde{n}^\ast_m$, 
$\Delta_m ( \max\{n^\ast_m, \widetilde{n}^\ast_m\} )$ 
is a borderline element.  
Furthermore, from the proof of Lemma \ref{lem:disagree}, 
we know that each arm only has at most one borderline element.  
Thus, we know that for each arm, 
if it does has borderline elements, 
every optimal arm pulling profile agrees on it, 
i.e., assign the same number of players to it.  
If it has one borderline element and some players' estimate on 
the optimal arm pulling profile has this borderline element, 
but some player's estimate does not, 
Algorithm \ref{algo:Consistency} identifies this borderline element out.  
If it has one borderline element and each player's estimate
on the optimal arm pulling profile has it, 
Algorithm \ref{algo:Consistency} does not identify 
this borderline element out, but it does not matter, 
as all players have a consensus on this arm already.  
Namely, Algorithm \ref{algo:Consistency} only identifies the 
borderline element that cause players do not have a consensus 
on an arm out.  
In the second phase, Algorithm \ref{algo:Consistency} removes 
these identified borderline elements.  
After this, all players have the same assignment of players to arms.   
But we only eliminate elements, leading to that the number of assigned 
workers is less than the total number of workers.   
To preserve the optimality, each player adds the same 
number of elements from the tie  as eliminated ones.  
Thus, finally, Algorithm \ref{algo:Consistency} achieves a consensus.  
This proof is then complete.  

\subsection{Proof of Theorem \ref{thm:online:regret}} 

The per round regret can is bounded by $K$.  
Hence, the total regret in the exploration phase 
is bounded by $T_0 K$.  
It takes $M$ round to reach the consensus.  
The total regret in this phase is bounded by $MK$.   
The expected regret in the committing phase is bounded by 
 $KT_\text{commit}$.  
The probability of that at least one player fails to have an accurate 
estimate of the optimal arm pulling profile 
is $ MK\delta_1 + MK\delta_2 
+K\sum_{m\in \mathcal{M}}\exp\left(
-\frac{T_0c_m}
{8}
\right)$.   
If at least one player fails to have an accurate 
estimate of the optimal arm pulling profile, 
players will not commit to an optimal arm pulling profile.  
The total expected regret is bounded by 
$
(T-T_0-M-T_\text{commit})K^2
[M\delta_1+M\delta_2+\sum_{m\in \mathcal{M}}
\exp\left(
\frac{-T_0c_m}
{8}
\right)
] 
$.  
Hence, the total regret in all phase is bounded by  
$
T_0K+MK+KT_\text{commit}+(T-T_0-M-T_\text{commit})K^2
[M\delta_1+M\delta_2+\sum_{m\in \mathcal{M}}
\exp\left(
\frac{-T_0c_m}
{8}
\right)
] 
$.
To make $\sum_{m\in \mathcal{M}}\exp\left(
\frac{-T_0c_m}
{8}
\right)
\leq \frac{M}{T}$, 
we only need to set 
\begin{align}
T_0 \geq \max_{m\in \mathcal{M}}\frac{8\ln T}{c_m}
\end{align}
Setting $\delta_1=\delta_2=\frac{1}{T}$ and according to the Theorem \ref{thm:estOptProfile}, we have 

\begin{align*}
T_0 
&= \max_{m\in \mathcal{M}}
\frac{
(2/c_m)
\ln{\delta_1^{-1}}
\ln{\delta_2^{-1}}
}
{
\left(
\sqrt{
0.5\ln{\delta_1^{-1}}
+{c_m}^{-1}
\ln{\delta_2^{-1}}
+(2+2\gamma)
(2c_m)^{-1/2}
\sqrt{\ln{\delta_1^{-1}}\ln{\delta_2^{-1}}}
}
- \sqrt{0.5
\ln{\delta_1^{-1}}}
- \sqrt{c_m^{-1}
\ln{\delta_2^{-1}}}
\right)^2
}\\
&= \max_{m\in \mathcal{M}}
\ln T
\left[
\frac{2/c_m}
{\left(
\sqrt{
0.5   +    c_m^{-1}   +    (2+2\gamma)  (2c_m)^{-1/2}
}
-\sqrt{0.5}
- c_m^{-1/2}
\right)^{2}
}
\right].
\end{align*}
Choose $T_0=
\max\{\max_{m\in \mathcal{M}}
\ln T
\left[
\frac{2/c_m}
{\left(
\sqrt{
0.5   +    c_m^{-1}   +    (2+2\gamma)  (2c_m)^{-1/2}
}
-\sqrt{0.5}
- c_m^{-1/2}
\right)^{2}
}
\right]
,\max_{m\in \mathcal{M}}
\frac{8\ln T}{c_m}
\}=C\ln T
$, 

we can get that the regret is 
$
CK\ln T+MK+KT_\text{commit}
+3MK^2
-\frac{3(C\ln T+M+T_\text{commit})MK^2}{T}
=
O(\ln T)
$.  
This proof is then complete.    

\end{document}